\documentclass[final]{cvpr}

\usepackage{times}
\usepackage{epsfig}
\usepackage{graphicx}
\usepackage{amsmath}
\usepackage{amssymb}
\usepackage{multirow,booktabs}
\usepackage{array}
\usepackage{xcolor}
\usepackage{appendix}
\usepackage[misc]{ifsym}  % for \Letter
\usepackage{mathtools, cuted}
\usepackage{capt-of}

\definecolor{citecolor}{RGB}{65,105,225}
\usepackage[pagebackref=true,breaklinks=true,letterpaper=true,colorlinks,citecolor=citecolor,bookmarks=false]{hyperref}

\def\cvprPaperID{xxxx} % *** Enter the CVPR Paper ID here
\def\confYear{Technical Report}
\begin{document}
\def\doubleColumn{1}
%%%%%%%%% TITLE
\title{Simulating Fluids in Real-World Still Images}

\author{
Siming Fan\textsuperscript{1}\thanks{The first two authors contribute equally to this work, and assert joint first authorship.},
~~~ Jingtan Piao\textsuperscript{1,2}\footnotemark[1],
~~~ Chen Qian\textsuperscript{1},
~~~ Kwan-Yee Lin\textsuperscript{1,2 \Letter},
~~~ Hongsheng Li\textsuperscript{2 \Letter} \\
\textsuperscript{1}SenseTime Research \\
\textsuperscript{2}CUHK-SenseTime Joint Laboratory, The Chinese University of Hong Kong\\
% \textsuperscript{3}School of CST, Xidian University \\ 
{\tt\small 1155116308@link.cuhk.edu.hk, \{fansiming,qianchen,linjunyi\}@sensetime.com, hsli@ee.cuhk.edu.hk}
}

\maketitle
%\pagestyle{empty}
%\thispagestyle{empty}
% \twocolumn[{%
%             \renewcommand\twocolumn[1][]{#1}%
%             \vspace{-1em}
%             \maketitle
%             % \vspace{-3em}
%             % \begin{center}

\begin{strip}
    \centering
    \vspace{-1.5cm}
	\includegraphics[width=\textwidth]{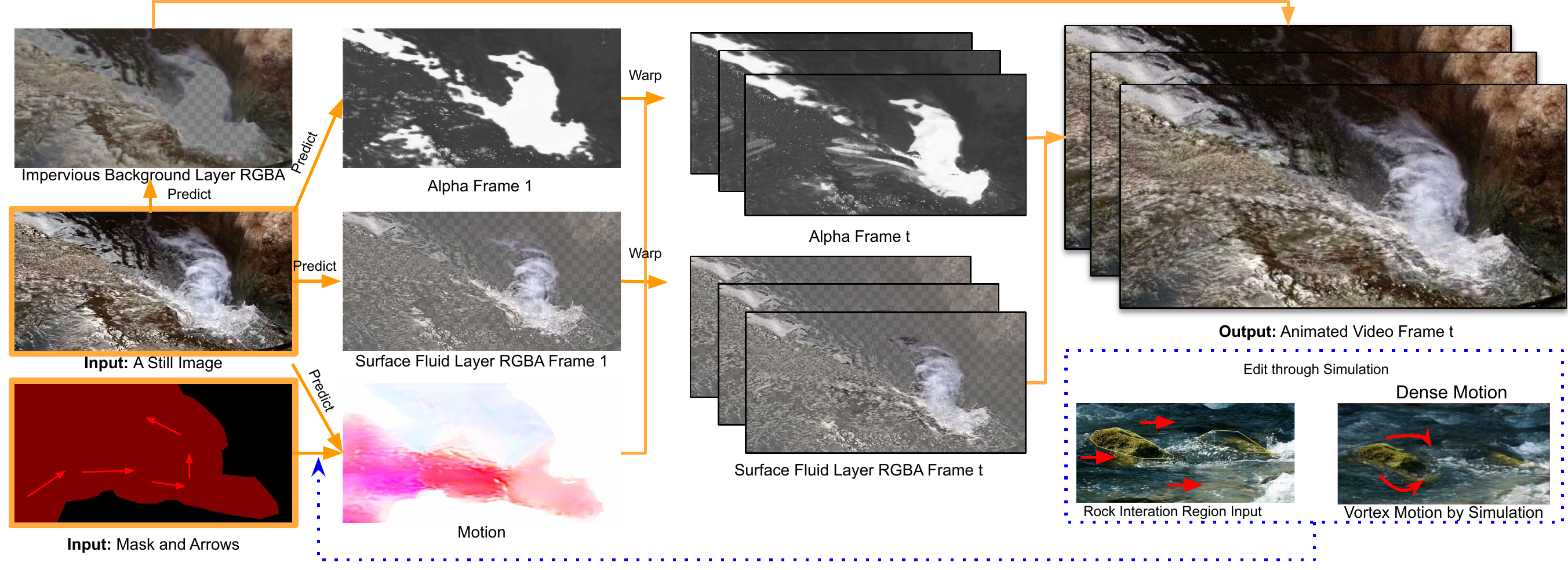}
	\captionof{figure}{\textbf{Overview.} Given a still image and a coarse hint of motion as inputs, our model estimates fluid motion to generate animating videos. To be able to represent complex scenes like transparent fluid shown in the figure, we propose to learn a single background RGBA layer and per-frame surface fluid RGBA layer to compose each frame of the final animated video ({\color{orange}solid arrows} indicate the data flow). Besides, a simulation-based motion editing method ({\color{blue}dot arrow} in the figure) is introduced to generate realistic effects like fluid-rock interaction, which cannot be easily captured by learning based method only. The edited motion direction is represented as red arrows.}
\end{strip}

%%%%%%%%% ABSTRACT
\vspace{-2mm}
\begin{abstract}

We have witnessed great progress in physical-based simulation and neural video generation for animating fluids. However, the two types of methods suffer from different drawbacks. The physical-based simulation methods are built upon manual-designed environments with specific materials, motion trajectories and textures, thus are only capable of animating particular fluids in synthetic scenarios. On the other hand, the neural video generation methods usually encode and warp the entire scene as a whole, which are generally not aware of the complex contents, such as transparency, collision and thin structures that frequently appear in real-world scenarios. In this work, we tackle the problem of real-world fluid animation from a still image. The key of our system is a surface-based layered representation deriving from video decomposition, where the scene is decoupled into a surface fluid layer and an impervious background layer with corresponding transparencies to characterize the composition of the two layers. The animated video can be produced by warping only the surface fluid layer according to the estimation of fluid motions and recombining it with the background. In addition, we introduce surface-only fluid simulation, a $2.5D$ fluid calculation version,  as a replacement for motion estimation. 
Specifically, we leverage the triangular mesh based on a monocular depth estimator to represent the fluid surface layer and simulate the motion in the physics-based framework with the inspiration of the classic theory of the hybrid Lagrangian-Eulerian method, along with a learnable network so as to adapt to complex real-world image textures. We demonstrate the effectiveness of the proposed system through comparison with existing methods in both standard objective metrics and subjective ranking scores. Extensive experiments not only indicate our method's competitive performance for common fluid scenes but also better robustness and reasonability under complex transparent fluid scenarios. 
Moreover, as the proposed surface-based layer representation and surface-only fluid simulation naturally disentangle the scene, interactive editing such as adding objects to the river and texture replacing could be easily achieved with realistic results. Code, model and dataset are publicly available\footnote{Project page: \url{https://slr-sfs.github.io/}\\
Code and model: \url{https://github.com/simon3dv/SLR-SFS}}.

\end{abstract}

\section{Introduction}

Given an image of a scene containing liquid regions such as streams, waterfalls and oceans, humans can imagine how the liquid in the still image would move. 
The problem of animating fluid from a single image has gradually become a blooming topic recently with a series of works  \cite{endoSA2019,holynski2021animating,mahapatra2021controllable,halperin2021endless,le2022animating}. Their general frameworks can be summarized as two parts: $(1)$ predicting/calculating the optical flow that indicates the movement of the liquid and $(2)$ synthesizing the future frame image based on the motion prediction/calculation. Although these methods achieve impressive visual effects on simple fluid regions, how to handle complex contents with challenging transparency, collisions, and thin structures in real-world scenes is still a challenging and open problem.

\begin{figure}[htbp]
\ifx\doubleColumn\undefined
	\centering
	\includegraphics[width=\textwidth]{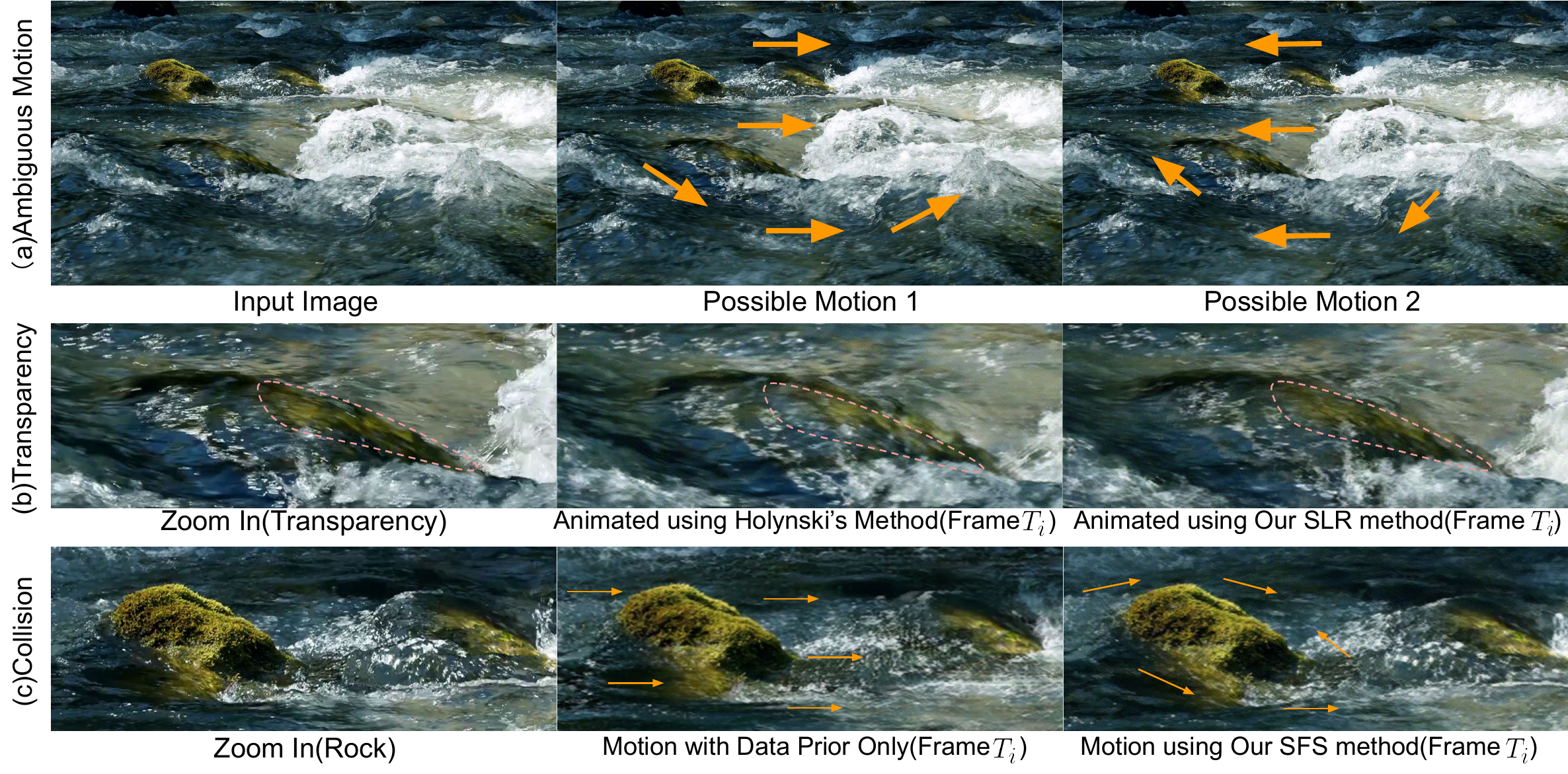}
\else
	\includegraphics[width=\linewidth]{fig/Motivation.pdf}
\fi
	\caption{\textbf{Motivation}. Given a still input image, (a)motion may be ambiguous; (b)animating with previous SOTA method \cite{holynski2021animating} causes background rocks beneath the liquid improperly deformed; (c)motion prediction method like \cite{holynski2021animating, mahapatra2021controllable} cannot well capture motion changes causing by the collisions with objects.}
	\label{fig:AmbiguousVelocity}
\end{figure}

The key to the limitation of previous work is that the generation process of videos is modeled in a global manner. Specifically, the subsequent frames are generated based on warping existing textures on input images \cite{chuang2005animating} or neural features \cite{endoSA2019}. Such operations regard the scene as a whole and warp all the contents in the scene together. For some complex transparent fluid cases, as shown in Figure~\ref{fig:AmbiguousVelocity}(b), where we can see through the water that the background rocks beneath the liquid, textures on those static objects are also improperly deformed and result in unnatural animations. To decompose the influence of motion on fluids and surrounding static objects, we propose a two-layer representation for scenes with fluids, namely \textbf{S}urface-based \textbf{L}ayered \textbf{R}epresentation (SLR), which models the motion of the surface fluid and the rest background separately. 
To achieve such decomposition from the still image during the inference,
we propose a temporal-based training schema, which fascinates the network to learn the decomposition with the help of several supervisions over spatial and temporal dimensions of running fluid videos. 

On the other hand, another major difficulty in animating a single fluid image is predicting the motion flow of the fluid. There exist two types of pipelines. 
The first category is the interactive motion prediction methods. 
They require sparse hints (\textit{ie}, labelling the velocity direction and relative amplitude)
on random pixels of the liquid region as additional input. 
The feature clustering model \cite{mahapatra2021controllable} or convolutional network \cite{holynski2021animating} is then used to form a dense motion field. As the precision depends on the level of motion details a user can provide, such methods are time-costing in complex fluid scenes. 
The second category is automatic methods \cite{endoSA2019}. These work regards the task as a domain transformation problem, and
use image-to-image translation techniques to predict the motion
 representing by $2$-channel optical flow.
Ambiguous velocity (e.g., a flat river surface can flow either to the left or to the right, as shown in Figure ~\ref{fig:AmbiguousVelocity}(a)) and tremendous changes of velocity in a period of time caused by the collisions with objects (as shown in Figure ~\ref{fig:AmbiguousVelocity}(c))
may pose great challenges to the above methods. 
To tackle the challenges, we 
introduce a simplified Navier-Stokes simulation into the motion estimation process, called {\textbf{S}}urface-only {\textbf{F}}luid {\textbf{S}}imulation (SFS). Specifically, the initial motion prediction of the liquid region is firstly calculated through sparse user labels and interpolating inside the whole region. In parallel, a 3D mesh is built based on a monocular depth estimator. Then, the N-S equation is solved on the mesh surfaces and based on the initial motion to obtain a simulated motion field in a short period of time. Finally, since the real thickness of the fluid is neglected during the simulation, a refinement step is expected to compensate for the motion offset at height. Thus,
a DNN-based motion translator is applied subsequently to obtain the refined motion field that adapts naturally to the input image.
With SFS, we can enforce precise controls on the movement of fluid. 

We can easily extend our fluid animation system to downstream editing tasks, thanks to the SLR and SFS designs in our framework. For example, we can augment objects in the fluid region or background region and simulate the interactions between the fluids and objects. We can also replace textures to different layers to create different scenes.

In summary, our work has three major contributions.
(a) We propose a new and learnable representation, surface-based layered representation, which decomposes the fluid and the static objects in the scene to better synthesize the fluid animating videos from a single fluid image. 
(b) We design a surface-only fluid simulation to model the evolution of the image fluids with better visual effects.
(c) Based on the proposed surface-based layered representation and surface-only fluid simulation, image editing with fluids is achieved with realistic and vivid results.
\section{Related Work}

This section will briefly introduce some of the prior works on neural video generation, fluid simulation, and motion prediction to give a picture of the related areas. 

\subsection{Video Generation}

In early studies, researchers focus on generating new video context by changing the attributes in which the recorded frames are played~
\cite{reinhard2001color,schodl2000video}. Traditional methods of video motion generation with movable substances are based on stochastic process analysis \cite{DBLP:conf/iccv/sunJF03} and color transfer \cite{reinhard2001color}. Given a reference video indicating the temporal changes of a scene, texture movement and sequential images can be generated using a patch aligned method and energy-based optimizing \cite{schodl2000video}. Furthermore, similar methods are applied to single image input \cite{chuang2005animating}. However, obvious artefacts can be detected, and the smoothness of the video is not guaranteed,  making the usage of such methods limited in typical types of scene samples.

In recent years, since the development of deep neural networks in the image generation field~\cite{pix2pix2017}, plenty of innovative efforts have been made to improve the synthesis quality of the images \cite{2016Image,2016Perceptual,2014Generative,CycleGAN2017,liu2018image}. Techniques such as generative adversarial training \cite{2014Generative}, perceptual loss \cite{johnson2016perceptual}, partial convolution \cite{liu2018image} and cycle-consistency \cite{CycleGAN2017} have been widely used as standard components. Built upon the success of image generation, several attempts have created the precedents of neural video generation. A few methods are developed to force the network to learn intermediate scene representation for predicting the future frames or interpolating key frames, once the input sequences and camera trajectory are given (note: as we focus on single image input, we skip the review of such multi-frames input trend). Given a single image as input, latent generative methods like~\cite{infinite_nature_2020} proposes an autoregressive model to generate infinite pixels forming sequential outputs. InfinityGAN~\cite{lin2021infinitygan} disentangles global appearances, local structures and textures with a structure synthesizer and texture synthesizer to generate images with spatial size and level of details not attainable before. \cite{prashnani2017phase} tries to use Fourier transforms in time variance showing the phase variations to represent the flow and generate moving videos. \cite{laffont2014transient} spends efforts in digging into the semantic meanings in images and generates videos correspondingly to achieve a natural result. Animating Landscape \cite{endoSA2019} decouples motion and appearance through learning intermediate flow fields and color transfer map. While, these methods are either focusing on spatial-wise consistence \cite{laffont2014transient}, or paying attention to coarse-level time-variance scene changes \cite{prashnani2017phase,endoSA2019}. None of them discusses or can handle the particularity of \textit{fluid} with fine detail motions. 

\subsection{Fluid Simulation}

\noindent\textbf{Physical Fluid Simulation.} Given initial motion, we need to predict the following motion based on the initial state. Traditional methods originated from computer graphics. Over the years, several efforts have been made to achieve accurate motion approximations on various types of fluids. 
Hybrid Lagrangian-Eulerian Methods \cite{bridson2015fluid,zhu2005animating,harlow1964particle} have successfully simulated the motion of incompressible fluids in a pre-defined 2D or 3D box environment, which solves pressure projection on Eulerian grid and advect on Lagrangian particles.To improve the convergence ability and flexibility on complex shapes, the material point method (MPM) \cite{bardenhagen2004generalized} has been popular and more subsequent methods \cite{kim2005flowfixer,kim2008wavelet} have been proposed to stabilize the convergence when iterating velocities. However, these methods are time-consuming in large scenes. To reduce the cost of simulating a massive range of fluid, methods have been proposed to simulate only a small range of fluid \cite{citro2017efficient}, or the surface of the fluid \cite{da2016surface,huang2021ships} to produce a faster approximation by avoiding high computational cost in 3D volumetric solvers. Although these works achieve amazing simulation effects, they are still limited to synthetic scenarios due to scene-specific modelling. 

\noindent\textbf{Animating Fluids with Data Prior.} Given a still image as animation target and a set of real-world water video candidates as animation bank, \cite{prashnani2017phase, okabe2018animating} propose to leverage video retrieve to find a/a group of videos from the bank that has similar content with target image to as the references. Then they transfer motion and appearance of the references to each region of interest on target image to form the water animation video. These methods open the door of real-world fluid animation with data prior in-the-wild. However, the seamless alignment of animated results with source input image can not be well guaranteed and requiring lots of manual efforts. With the development of deep learning, recent works \cite{holynski2021animating,mahapatra2021controllable} propose to automatically predict each frame of motion and appearance via learning from reconstruction loss with real world videos for training.
\cite{holynski2021animating} first simplifies the motion estimation part as a single frame motion prediction via motion Eulerian integration and then proposes a deep feature warping technique to narrow down the size of blank areas caused by warping. Finally, it adaptively blends forward features and backward features through a learnable parameter Z to generate animated fluid video. Based on~\cite{holynski2021animating,mahapatra2021controllable} proposes to regress fluid motion conditioned to user's sparse guidance and generate paired training data through motion speed clustering. They further proposes to use multi-scale representation to capture different fluid speed in different resolution. Although these methods take a step further to real-world scenes with impressive results, it is hard for them to handle complex context relations over fluids due to the single representation to the entire scene that regarding different objects and textures as equal. 

Unlike previous work, our system marries the advantages of both physical simulation and learning-based pipelines. For textures, to avoid interplay between fluids and surroundings, we force the network to decouple the scene into a new two-layer representation \textbf{SLR}, that includes a static background and time-varies surface fluid layers. Meanwhile, the network can hallucinate reasonable and photo-realistic textures in vacated regions thanks to the large-scale data prior. For motion, to skirt scene-specific modelling while keeping the physic reasonability, we propose a three-step motion prediction. We first use sparse labels from the human interface to generate the initial rough motion and then calculate reasonable evolution in a short period by introducing a $2.5$D surface simulation, {\textbf{SFS}}. Later, a smooth motion translator is used to refine the motion trend to better fit real-world examples with data prior. Moreover, as a consequence, we can generate natural fluids animations with the network training on plenty of real-world examples and flexibly edit the main liquid region and boundary condition to create various visual effects. 

\subsection{Motion Prediction}

The acquisition of the initial velocity field, which is always described as the optical flow \cite{ilg2017flownet} of two sequential images, is difficult to predict from one single image due to ambiguous velocity. Originated from continuous modelling of the image colors with respect to coordinates, motion is predicted using the estimated differential of colors \cite{le2005dense}. Since convolutional neural networks are widely used in various tasks, deep learning-based methods \cite{ilg2017flownet,sun2018pwc} give more stable results on complex images. Recent works turn to view the task as an image-to-image translation task. Conditional Generative Adversarial Networks~\cite{mirza2014conditional} and Variational Autoencoder \cite{pu2016variational} and similar methods have been proposed to fulfil the task of transferring image to optical flow map, which try to model the process as a parameter-based probability matches and generate a probability distribution resembling the target flow, given the image input as the condition. Some self-supervised methods \cite{liu2019selflow} automatically learn a flow from a given video with the help of occluded boundaries and semantic masks to solve discrete color changes at the boundaries of objects in complex videos. Some work \cite{walker2015dense,zhan2019self} takes a single image as input and outputs an optical flow map using U-Net shaped network to capture different scale movements in the whole image. The direction of movements is divided into eight discrete directions, and for each, an absolute value shows fast or slow, and the probability is predicted to simplify the training. There are also some interactive motion prediction systems \cite{mahapatra2021controllable,zhan2019self,le2022animating}
designed to generate a more accurate motion with as little user guidance as possible.
\section{Method}

Given a real-world image that contains liquid regions, our goal is to generate a video clip that animates vivid flowing of the fluid. To this goal, a system is required to $(1)$ represent complex context relations between fluids and surrounding environments;
and $(2)$ to estimate motion fields from single image input with reasonability and efficiency. Thus, we propose a novel system with two main designs that satisfies the above requirements. For texture, in contrast with previous methods that regard all elements in the scene as an entirety, our system learns a surface-based layered representation (SLR) (Sec.~\ref{sec-slr}) that decomposes the scene into surface fluid layer and impervious background layer. Since the texture characteristics of surface fluid and rest background are assigned into different layers, such design could suppress the negative influence of motion flows between fluids and surroundings (e.g., improper texture warping, as shown in Figure \ref{fig:AmbiguousVelocity}(b)). As for motion, the system provides a three-step motion estimation alternative with introducing a surface-only fluid simulation (SFS) into the motion field prediction (Sec.~\ref{sec-sfs}). The motion prediction steps combine the advantage of both physic- and learning-based motion estimation pipelines, thus ensuring the physic reasonability and meanwhile skirting the scene-specific modelling problem. We will detail each component in the following subsections.

\subsection{Learning of Decomposition of Images}\label{sec-slr}

\begin{figure*}[!t]
	\centering
	\includegraphics[width=\textwidth]{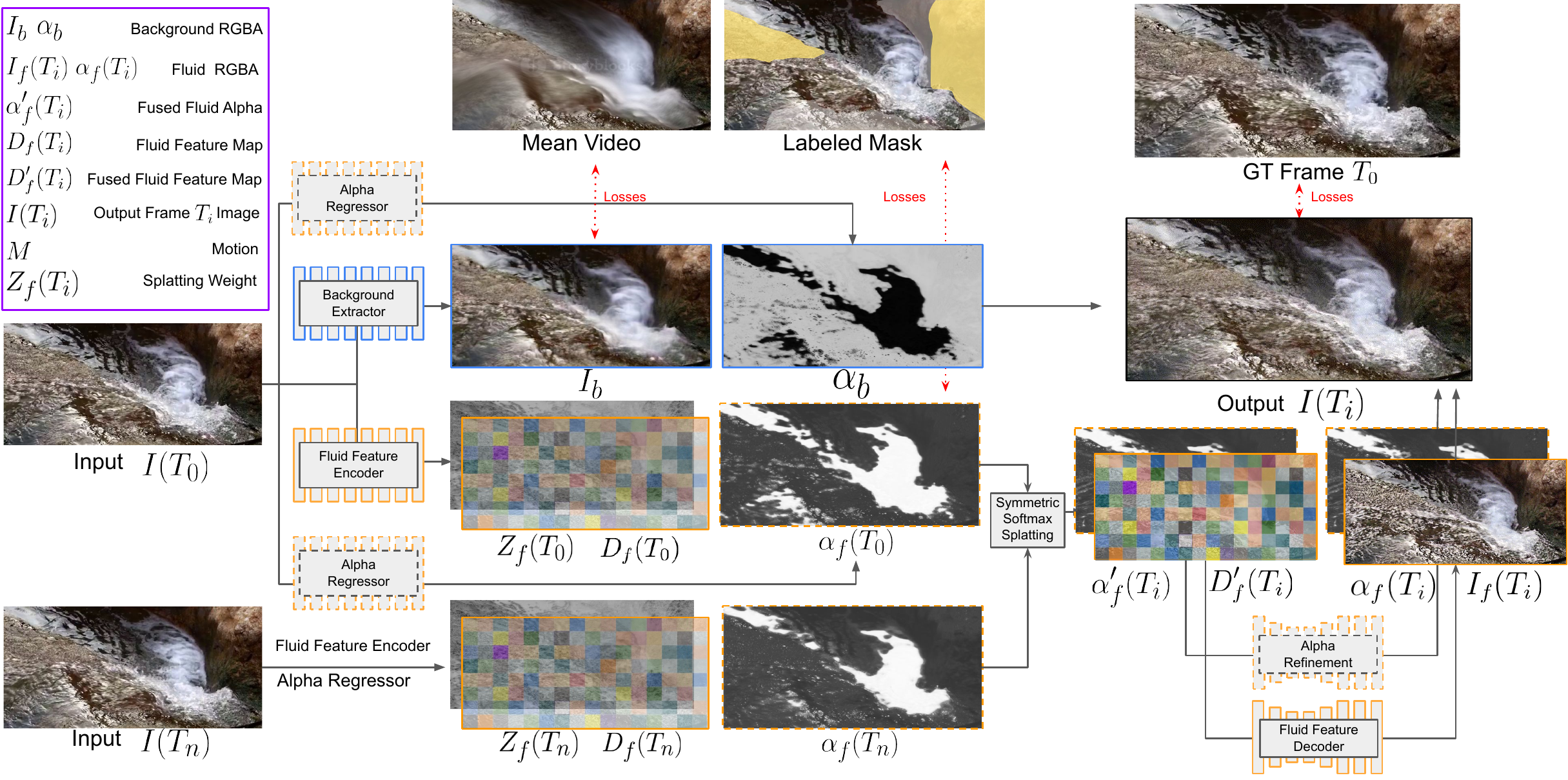}
	\caption{\textbf{Training and testing pipeline}. Background layer and background alpha is predicted only at the first frame. Both fluid features and fluid alpha are predicted at the first frame and the last frame, and then symmetric splatted and blended using softmax at frame t. A decoder is used to decode fluid features into image and refine alpha. For frame 1, motion is calculated following pipeline Figure ~\ref{fig:motion_pipeline}, and for frame t, motion is then calculated through Euler integration. When testing, $T_{n}$ is replaced by $T_{0}$ to create a cyclic video.}
	\label{fig:pipeline*}
\end{figure*}

In this subsection, we first introduce our proposed Surface-based Layered Representation (SLR) with formula. Then, we show how to
obtain the animated video through SLR. Finally, we present how to learn SLR with a neural network under encoder-decoder architecture (illustrated in Figure~\ref{fig:pipeline*}), and what kind of challenges will we meet during optimizing the network since there is no oracle ground-truth supervision for fluid scene decomposition,  as well as how we solve the problems through several loss and training strategy designs.

\noindent\textbf{Surface-based Layered Representation.} Given a still fluid image $I(T_{0})$ as input, a series of RGB frames $\textbf{I}=\{ I(T_{0}), \dots, I(T_{n}) \}$ are expected outputs to synthesize the time-space variation of the scene with running fluids. As we would like to handle the complex context of the fluid scene through disentangling the fluid motion and the still objects beneath/interacting with the surface fluid, we propose to learn a novel intermediate representation whose output is an impervious background layer image{\footnote{The word ``impervious" is not rigorous, as background layer is not only formed from static object but also non-surface fluid which contains air molecules in real environment. However, we assume the fluid is incompressible, and thus a little bit abuse the word ``impervious background" to indicate the non-surface fluid regions.}}~$I_{b}$, a time sequence of surface fluid layer 
images~$\textbf{I}_{f}=\{I_{f}(T_0),\dots,I_{f}(T_n)\}$ and their transparency factors $\alpha_{b}$, $\boldsymbol {\alpha}_{f} =\{\alpha_{f}(T_0),\dots,\alpha_{f}(T_n)\}$, where $\alpha_{*}\in \left[0,1\right]$. The transparency factors indicate the contribution of color on each layer to the final image of each time.
In practice, we utilize a convolution network with encoder-decoder architecture to output the \textit{ simplification} of the intermediate representation: $F_\theta:(I(T_{0}),I(T_{n}),\mathbf{M})\rightarrow(I_{b}, {I}_{f}(T_{0}),{I}_{f}(T_{n}), \alpha_{b}, {\alpha}_{f}(T_{0}),{\alpha}_{f}(T_{n}))$, where the network only outputs the estimated \textit{first} and \textit{last} frame of $\mathbf{I}_{f}$ as well as ${\boldsymbol {\alpha}_{f}}$.
$\mathbf{M}$ is the estimated motion fields of $T_0$ to $T_1$ from single input $I(T_{0})$. We will detail the synthesis of $\mathbf{M}$ in Sec.~\ref{sec-sfs}, let's assume it already exists for a while. 

We expect the background to be identical across all time, while the fluid surface deforms with realistic motions. To this end, the $F_\theta$ is enforced to {\textit{automatically}} map the original texture of non-liquid and static regions as well as the reasonable texture of under-liquid surface regions to the background layer(detailed in later subsection). As for fluid surface learning, recurrent estimation for all frames is a straightforward solution that could be tailored from previous video generation \cite{2019video}. While such design will raise distortion as time goes on due to the accumulated errors and hard to converge, as also demonstrated in~\cite{holynski2021animating}. In contrast, 
We use the linear combination of the first and last frames to construct the images at intermediate frames. This choice not only transforms the problem as interpolation rather than extrapolation, which is easier to converge, but also is able to generate an endless cyclic video rather than a video in a fixed interval. In detail, analogous to pixel warping that utilized in \cite{endoSA2019}, we regard each frame at a specific time $T_{i}$ as an interpolation of the previous frame $T_{0}$ and subsequent frame $T_{n}$ during training, and replacing $T_{n}$ to $T_{0}$ during testing for creating a cyclic video.

With a warping on the start frame $I_{f}(T_{0})$ and the final frame $I_{f}(T_{n})$, we can obtain two \textit{partial} fluid surface images, denoted as $I_{f}(T_{0}\to T_{i})$ and $I_{f}(T_{n}\to T_{i})$, that is pixel-aligned with $I_{f}(T_{i})$. With the motion field  $\mathbf{M}$, the warping can be viewed as
\begin{align}
	I_{f}(T_{0}\to T_{i}) &= I_{f}(T_{0}) \circ M_{i}(x, y)\nonumber\\
	I_{f}(T_{n}\to T_{i}) &= I_{f}(T_{n}) \circ M_{n-i}^{-1} (x, y)
\end{align}
where $\circ$ means the look-up operation by indexing the color on the updated position warped by motion field $M$, and $M_{i}(x,y) = (x+\int\mathbf{M}_{x}dt, y+\int\mathbf{M}_{y}dt)$. $\mathbf{M}_{x}$ and $\mathbf{M}_{y}$ indicates the vertical and horizontal velocity respectively. The position of $(x,y)$ from $T_{0}$ to $T_{i}$ is calculated iteratively from the first frame to current frame $T_i$. To keep smoothness and continuity in texture, we further use softmax splatting \cite{niklaus2020softmax} with learnable composition factors $Z(T_{0})$ and $Z(T_{n})$ to determine the contribution of overlapped pixels from $I_f(T_0)$ and $I_f(T_n)$ to the transformation, as also demonstrated in \cite{holynski2021animating}. The vacant pixels are left blank in this stage. Similar operations are adopted for the transparency map sequence $\alpha_{f}(T_{0}\to T_{i})$ and $\alpha_{f}(T_{n}\to T_{i})$. 

In practice, along with rgb color of the images of fluid layer (\textit{i.e.}, $I_{f}(T_{0})$ and $I_{f}(T_{n})$ ) {\footnote{``fluid layer" equals to ``surface fluid layer" in the rest of paper for convenience, unless otherwise specified.}}at time~$T_0$ and $T_n$, we also concatenate deep features extracted from a \textit{fluid} feature encoder with ResNet-based structure, to form $D_{f}(T_{0})$ and $D_{f}(T_{n})$, and the warping versions of them are named as $D_{f}(T_{0}\to T_{i})$ and $D_{f}(T_{n}\to T_{i})$. The composition factors $Z(T_{0})$ and $Z(T_{n})$ are also predicted to interpolate deep fluid features at frame $T_i$ between $D_{f}(T_{0}\to T_{i})$ and $D_{f}(T_{n}\to T_{i})$. The fused feature of target frame, $D'_{f}(T_{i})$ is then obtained through a linear combination as follows:
\ifx\doubleColumn\undefined
\begin{align}
	D'_{f}(T_{i}) &= \frac{	D_{f}(T_{0}\to T_{i})e^{Z(T_{0})}(T_{n}-T_{i}) +
				D_{f}(T_{n}\to T_{i})e^{Z(T_{n})}(T_{i}-T_{0})}
			{e^{Z(T_{0})}(T_{n}-T_{i}) + e^{Z(T_{n})}(T_{i}-T_{0})} \nonumber\\
	\alpha'_{f}(T_{i}) &= \frac{\alpha_{f}(T_{0}\to T_{i})e^{Z(T_{0})}(T_{n}-T_{i}) +
				    \alpha_{f}(T_{n}\to T_{i})e^{Z(T_{n})}(T_{i}-T_{0})}
			{e^{Z(T_{0})}(T_{n}-T_{i}) + e^{Z(T_{n})}(T_{i}-T_{0})} 
\end{align}
\else
\begin{scriptsize}
\begin{align}
	D'_{f}(T_{i}) &= \frac{	D_{f}(T_{0}\!\to\! T_{i})e^{Z(T_{0})}(T_{n}\!-\!T_{i}) \!+\!
				D_{f}(T_{n}\!\to\! T_{i})e^{Z(T_{n})}(T_{i}\!-\!T_{0})}
			{e^{Z(T_{0})}(T_{n}\!-\!T_{i}) \!+\! e^{Z(T_{n})}(T_{i}\!-\!T_{0})} \nonumber\\
	\alpha'_{f}(T_{i}) &= \frac{\alpha_{f}(T_{0}\!\to\! T_{i})e^{Z(T_{0})}(T_{n}\!-\!T_{i}) \!+\!
				    \alpha_{f}(T_{n}\!\to\! T_{i})e^{Z(T_{n})}(T_{i}\!-\!T_{0})}
			{e^{Z(T_{0})}(T_{n}\!-\!T_{i}) \!+\! e^{Z(T_{n})}(T_{i}\!-\!T_{0})} 
\end{align}
\end{scriptsize}
\fi
The fused $\alpha'_f$ is composed in the similar way. Then the fused fluid feature $D'_{f}(T_{i})$ is decoded by a fluid feature decoder to get final \textit{completed} $I_{f}(T_{i})$, and the \textit{completed} $\alpha_{f}(T_{i})$ is obtained through a alpha refinement network applied on $\alpha'_{f}(T_{i})$. Note that although the warping result $D_{f}(T_{0}\to T_{i})$ and $D_{f}(T_{n}\to T_{i})$ can cover most vacant pixels, the fused feature $D'_{f}(T_{i})$ is still a partial result with holes.
To obtain the completed fluid surface images and transparency maps, the fluid feature decoder and alpha refinement network are built with partial convolutions to hallucinate the rest vacated regions from the context.

\noindent\textbf{Animating Fluids with SLR.} 
Once we get the final surface fluid layer $I_{f}(T_{i})$ and transparency $\alpha_{f}(T_{i})$, the reconstructed final image $I(T_{i})$ can be acquired by adding the impervious background layer $I_b$ back to the surface fluid layer $I_{f}(T_{i})$ with $\alpha_b$. Here we use separate $\alpha$ channels for background and fluid transparency predictions. This helps to stabilize the training of the decomposition. The core reason is that when a single fluid transparency $\alpha_{f}$ is warped according to the motion flow,  it'll leave a blank region on the boundary, which leads the gradient hard to backpropagate. While with the two-channel implementation, there still is a $\alpha_{b}$ on that vacated regions to fascinate the network refining the texture. In the formula, the process can be written as

\begin{equation}
	I(T_{i}) = \frac{\alpha_{f}(T_{i})I_{f}(T_{i}) + \alpha_{b}I_{b}}
			{\alpha_{f}(T_{i}) + \alpha_{b}}
\end{equation}

For corresponding network architectures to achieve the above targets, as shown in Fig~\ref{fig:pipeline*},the encoder is used to extract background texture and $\alpha$ channel as well as surface fluid layers' features and its $\alpha$, the decoder is used to refine the fused fluid layer that combines the layers from the first and last frames to obtain the complete surface fluid layer, and the final animated image is reconstructed by compositing background and surface fluid layers. 

\noindent\textbf{Optimizing the SLR.}\label{opt}
We have introduced the SLR and the relation among each components in the previous subsections. Now, we describe how to fascinate the network to learn SLR with leveraging the powerful temporal and texture priors from large-scale in-the-wild videos at hand.

 The training process is divided into three parts, in which, we separately train $(1)$ the surface fluid branch containing the encoder that predicts the fluid feature layer from image input and the decoder that reconstructs the warped fluid layer, $(2)$ the background branch that predicts the background layer from the image input, and $(3)$ jointly train them together with $\alpha$ learning and surface fluid/ background branches finetuning. For each training step, the paired training data $<(I(T_0),I(T_n)), I(T_i)>$ are randomly picked among $60$-frames sequence without constraint on interval.
 We use dense optical flow estimation network \cite{ilg2017flownet} to extract reasonable movement of each video as pseudo motion ground-truth, and to ensure smoothness, we average the flow with window size of $30$ frames. 
For the first stage, we train the surface fluid branch with reconstruction losses at hand:

\begin{align}
	\mathcal{L}_{\text{image}} = &\vert I(T_{i}) - I_{gt}(T_{i})\vert + \nonumber\\
		\lambda_{0}&\Vert\text{VGG}(I(T_{i})) - \text{VGG}(I_{gt}(T_{i}))\Vert +\nonumber\\ 
		\lambda_{1}&Disc(I(T_{i}))
\end{align}

where $Disc$ denotes discriminative loss using a spectral normalized network updating simultaneously to distinguish the facticity of the output fluid image.  The target of this stage is to enable the decoder to have the ability to memorize the texture of the fluids and hallucinate the texture in blank areas caused by warping. The background is viewed as black in this stage. 

For the second stage, we train the background branch with warm-up supervision. 'Averaged image' of the whole sequence is used as guidance. The general idea behind this implementation is that for static textures in the video, averaging operation does not decline their quality. While for flowing textures, averaging will lead to blurring as well as mean static of texture,  which is reasonable for under-surface liquid, as shown in Figure \ref{fig:pipeline*}. The loss could be written as:

\begin{equation}
	\mathcal{L}_{\text{bg}} = \vert I_{b} - \frac{1}{n}\sum_{i}I(T_{i})\vert
\end{equation}

For the last stage, all the network components are trained together to update the $\alpha$ channels that composite the two layers, as well as to fine-tune other parts of the network.
 Despite the loss mentioned in the previous stages, we also make some restrictions on the predicted $\alpha$ that share a similar spirit with~\cite{lu2021omnimatte}. Considering $\mathbf{I}_{gt}(T_i)$ cannot be well aligned with $I(T_i)$ as motion is a pseudo ground truth, only supervised with image reconstruction loss will mislead alpha regressor to incorrectly prediction for scenes under complex motion variation. For example, splashes will always appear in the first frame then disappear in the next frame or do not exist in the first frame but appear in the next frame under a fluid collision scenario, as shown in the white splashes region in Figure \ref{AlphaLearning}(d).
Such variation is hard to infer from previous source frames. For this reason, we constrain  $\alpha$ with two loss terms to form  $\mathcal{L}_{\alpha}$ as follows: 

\begin{align}
	\mathcal{L}_{\alpha} = &\vert \left(\frac{\alpha_f(T_{i})}{\alpha_f(T_{i})+ \alpha_b}  -  \alpha_{label}\right)\odot R_{M>\gamma}\vert + \nonumber \\ &\vert (\alpha_f'(T_{i}) - \alpha_f(T_{i})) \odot R_{valid} \vert \nonumber\\ \label{alpha-cons}
\end{align}

The first term is a L1 Loss under moving region $R_{M>\gamma}$ to provide a hint for alpha learning, where $\gamma$ is a hyper-parameter that specifies the boundary of moving pixels and static pixels. $\odot$ is an element-wise product,  $\alpha_{label}$ is generated with our newly labelled mask in the solid region and $\frac{\alpha_f(T_{i})}{\alpha_f(T_{i})+ \alpha_b}$ is called the composited alpha{\footnote{It is abbreviated as ``alpha" in the later experiment section for convenience,  unless otherwise specified.}} .
The second term is a temporal-wise $\alpha$ consistency loss, which is applied between warped partial $\alpha'_f$ and its refinement complete result $\alpha_f$ under valid non-hole pixel regions in the target image .
 Then, the final loss terms in this stage is:
\begin{align}
    	\mathcal{L} &=	\mathcal{L}_{\text{image}} +
			\lambda_{\text{bg}}\mathcal{L}_{\text{bg}} +
			\lambda_{\alpha}\mathcal{L}_{\alpha}
\end{align}

\subsection{Motion Field Estimation}\label{sec-sfs}

\ifx\doubleColumn\undefined
\begin{figure}[htbp]
	\centering
	\includegraphics[width=\textwidth]{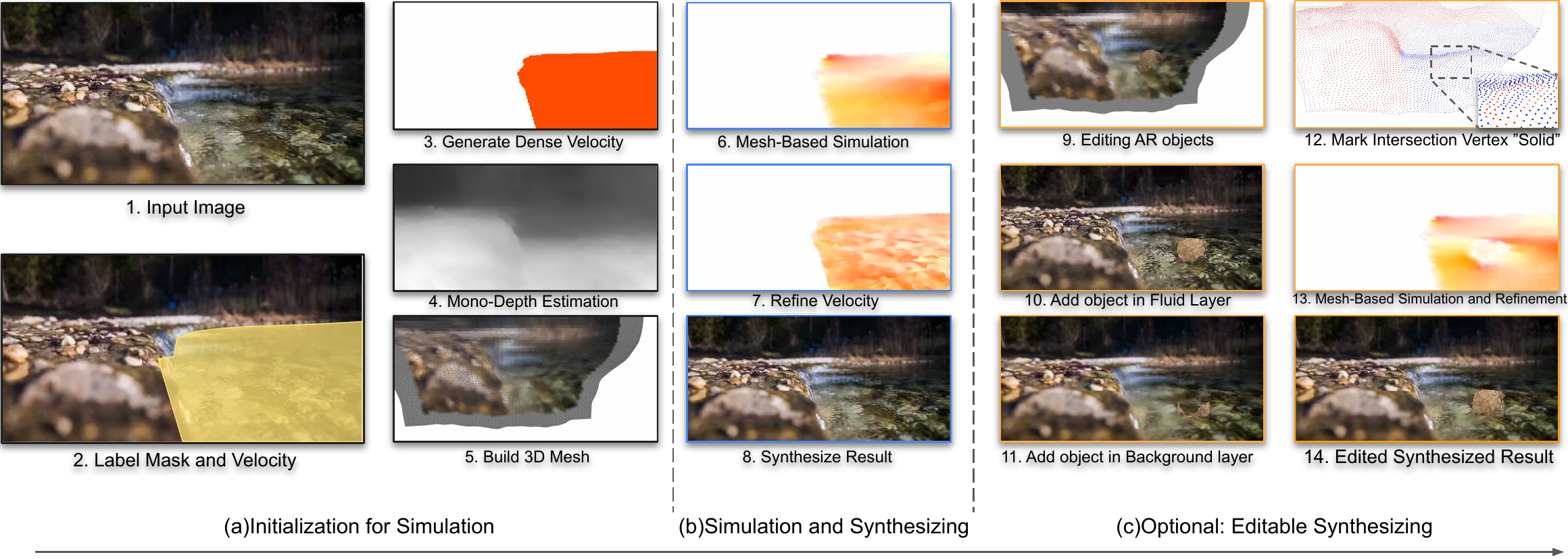}
	\caption{\textbf{Motion calculation pipeline}. Stage (a) is initialization of fluid region, rock region, motion, and Mesh. Stage (a,b) combine to form our 2.5D simulation and animation pipeline, where (4-5) is not included if using our 2D simulation method. Stage (a,c) combine to form our editable simulation and synthesizing pipeline, where the user can edit 3D objects and create corresponding motion.}
	\label{fig:motion_pipeline}
\end{figure}
\else
\begin{figure*}[thbp]
	\includegraphics[width=\textwidth]{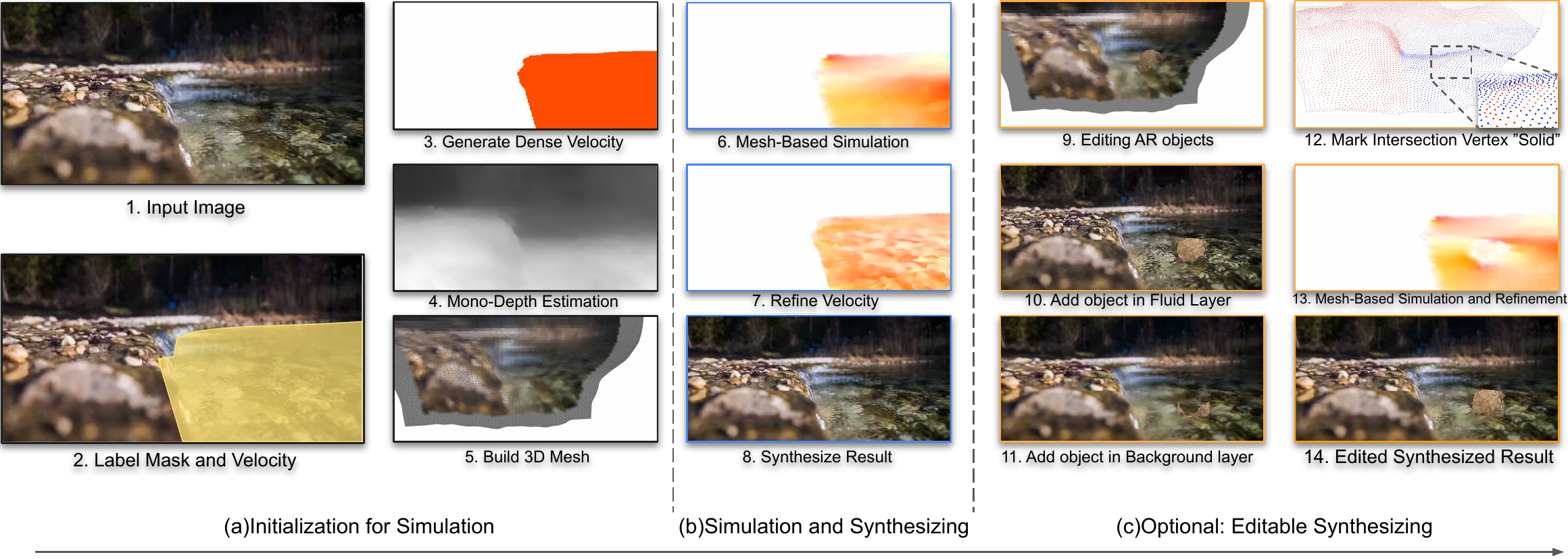}
	\caption{\textbf{Motion calculation pipeline}. Stage (a) is initialization of fluid region, rock region, motion, and Mesh. Stage (a,b) combine to form our 2.5D simulation and animation pipeline, where (4-5) is not included if using our 2D simulation method. Stage (a,c) combine to form our editable simulation and synthesizing pipeline, where the user can edit 3D objects and create corresponding motion.}
	\label{fig:motion_pipeline}
\end{figure*}
\fi

In the previous section, we have described the core necessaries to learn an SLR but skipped the synthesis process of one of its input, motion fields $\mathbf{M}$. In this section, we dive into the discussion of $\mathbf{M}$.

During training, the motion of each image frame can be generated with a conventional optical flow estimator \cite{ilg2017flownet} applied to the video sequence. While for testing, as the input is only a single image, we expect a module that could synthesise reasonable motion from the observed scene. Intuitively, we could directly adapt previous achievements: interactive labelling from users \cite{zhan2019self}, image to image translation learning~\cite{pix2pix2017} or physical-based simulation. However, the motion precision of interactive labelling depends on the density and partition affordance of user input. Image-to-image translation learning may not handle the ambiguous and choppy velocity mentioned in Figure \ref{fig:AmbiguousVelocity}. And physical-based methods are usually case-specific modelling and will give the wrong prediction 
if the initial state is undetermined.

Thus, we propose a motion estimation process that separates the motion prediction into three parts and stands on the shoulder of both physical-and learning-based pipelines for fluid simulation. First, interactive labelling of the sparse velocity of several pixels on the image is required to guide the preferred motion directions of $I(T_0)$, along with a fusion with weights calculated through Gaussian blur of distance maps to build a dense velocity initial map. 
Second, a surface-only fluid simulation is calculated based on the initial motion and estimated scene depth to simulate a following convincing evolution of the liquid. Note that this step only calculates in a short period, not for all frames. Last, a DNN-based motion translator under a pixel-to-pixel translation framework is utilized to refine the flow to a more realistic and natural motion. This translator could be understood as a compensation for detailed motion that is harmonic with a fluid texture. The pipeline is illustrated in Figure~\ref{fig:motion_pipeline}.

\noindent\textbf{Initial Motion from Interactive Sparse Labeling.} To obtain a determined as well as editable motion direction, we choose an interactive alternative to generate initial flow for the image $I(T_0)$  with following the work \cite{mahapatra2021controllable}. With no more than $10$ discrete notations on images, and a fixed segmentation of the liquid region, we use nearest-neighbors-averaging to give all the pixels that inside the liquid region an initial velocity. To be precise, the velocity at pixel $i,j$ is a exponential averaging of the neighbors as
\begin{equation}
	v_{i,j} = \frac	{\sum_{k} V_{k}e^{-d(k,ij)^{2}/\sigma^{2}}}
			{\sum_{k} e^{-d(k,ij)^{2}/\sigma^{2}}}
\end{equation}
where $v_{i,j}$ is the pixel inside the fluid region. $V_{k}$ is the $k$-th labeled velocity. $d(k,ij)$ is the Euclidean distance on image between the pixel (i,j) and the $k$-th labeled position. $\sigma$ is a hyper parameter proportional to image size.

\noindent\textbf{Surface-only Fluid Simulation.} After the initial state is estimated, we tailor the traditional graphics pipeline with discretization and linear equation to simulate the follow-up evolution of the liquid. Intuitively, we can directly apply simulation on 2D grids that align with image pixels by initializing each pixel as an Eulerian grid and several Lagrangian particles near each pixel. However, since such perspective projection of the 3D scene may cause severe distortion when direct manipulating the motion in 2D, as shown in Figure~\ref{fig:SimulationAblation}, we thus turn to use a mono-depth estimation network \cite{li2018megadepth} to estimate the 3D surface of the liquid. After the 3D surface is constructed inside the region of the liquid, a mesh surface-based velocity is calculated, with the restriction of projected velocity :   
\begin{align}
	\frac{d}{dt}\begin{bmatrix}
		u \\ v
	\end{bmatrix} &= \frac{d}{dt}\left(\begin{bmatrix}
		fx & 0 & cx \\
		0 & fy & cy
	\end{bmatrix}\begin{bmatrix}
		x/z \\ y/z 
	\end{bmatrix}\right) \nonumber \\
	&= \begin{bmatrix}
		fx & 0 \\
		0 & fy
	\end{bmatrix}\begin{bmatrix}
		1/z & 0 & -x/z^{2} \\
		0 & 1/z & -y/z^{2}
	\end{bmatrix}\begin{bmatrix}
		\frac{dx}{dt} \\
		\frac{dy}{dt} \\
		\frac{dz}{dt}
	\end{bmatrix}
\end{align}

Where $fx,fy,cx,cy$ is the camera intrinsic parameters, $u,v$ is the velocity on the projected image and $x,y,z$ the 3D position in the scene. We only list the final formula here. Please refer to Appendix~\ref{ref-formula} for a precise formula derivation. 

Since we have modelled the liquid as a surface, we replace the traditional Laplacian operator with the Laplacian-Beltrami operator on the mesh and use Mixed-Euler-Lagrange approach~\cite{zhu2005animating} implemented in Taichi~\cite{hu2019taichi} to simulate the consequent motion of each vertex. The NS-Equation could be written as:

\begin{align}
	(\frac{\partial}{\partial t} + v\cdot\nabla)v &= -\frac{1}{\rho}p + g\nonumber\\
	\nabla\cdot v &= 0
\end{align}

Where $v$ is the 3D velocity parallel to the fluid surface, $g$ is external forces that are represented by gravity in scenes such as waterfalls, $\rho$ is the density, $p$ is the pressure which is solved as one term by forcing the velocity to be divergence-free. $\nabla$ is calculated as Manifold-Differential on the mesh surface. Detailed formula derivation can be seen in the Appendix~\ref{ref-formula}. The particle binding on the surface is randomly sampled according to the surface area. When a particle is moved outside the mesh surface, we use the nearest projection forcing the position to fall onto the mesh. For the boundary of the mesh, we calculate the velocity of the boundary triangle, judging whether velocity on the edge boundary is inward or outward. For outward edges, the particle is abandoned after moving out of the liquid region. For inward region, a particle source is built to maintain the overall number of particles to be roughly stable. 

So far, with the texture of the surface fluid layer binding to each vertex, we can warp the fluid layer with a soft rasterizer to simulate the flows. In this manner, we can try an arbitrary flow initial state, and after a period of simulation, a stable velocity can be calculated to approximate the real-world flow. We call above process as Surface-only Fluid Simulation (SFS).

\begin{figure}[!htbp]
	\centering
	\includegraphics[width=\linewidth]{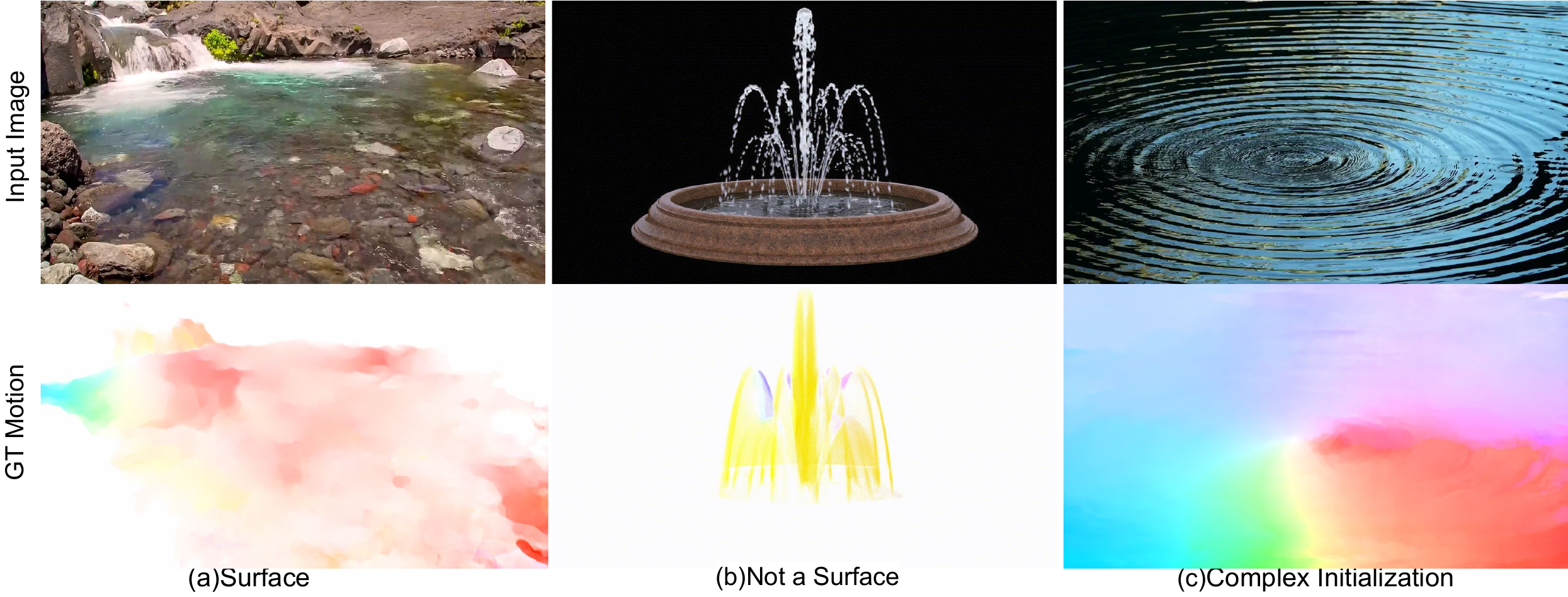}
	\caption{\textbf{Examples for Different Fluid Scenes}. (a) The surface thickness of the waterfall and river are gauzy. (b) The fountain cannot be considered a surface. (c) Fluid with vortex contains complex motion.}
	\label{fig:SimulationCannotHandle}
\end{figure}
\noindent\textbf{Smooth Motion Translator.} With the help of above $2.5D$ fluid simulation,  we can generate a time-variant velocity during the sequence. 
However, the simulated velocity is still a rough trend since it only works on the fluid surface without considering thickness variation of fluids. Thus, the SFS can work well in cases like Figure \ref{fig:SimulationCannotHandle}(a), it fails in cases with distinct depth measurement differences, as shown in the Figure \ref{fig:SimulationCannotHandle}(b) due to the lack of height information during calculation. Also, It is fragile to cases with complex initial state as shown Figure \ref{fig:SimulationCannotHandle}(c) due to error-prone/time-consuming interactive initial motion labeling for such cases. Thus, we need extra scheme to
compensate or rectify for detailed motion that is harmonic with the fluid textures.
To this end, we use a convolution-based network to transfer the simulated motion into a more detailed and reasonable one, that is pixel-aligned with image textures. The network is trained as a traditional style-transfer task, with a L2 loss between output velocity and ground truth. An external patch discriminator is used to refine the pattern of the velocity distribution. All the settings are similar to the image translation task pipeline \cite{pix2pix2017}. 

\section{Experiments}

In this section, we present experiments and detail analysis to
demonstrate the contributions of our method both quantitatively and qualitatively. We first show comparison with state-of-the-art learning-based fluid animation method on both Holynski and self-collected datasets to discuss the effectiveness of proposed Surface-based Layered Representation (Sec.~\ref{exp-slr}). Then, we discuss the influence of different motion field estimations to fluid animation, including our proposed Surface-based Fluid Simulation, to the quality of fluid animations (Sec.~\ref{exp-sfs}). Finally, we show some scene editing applications to give a brief exploration of the byproducts of our framework. {\textbf{For all the cases shown in this paper, please refer to our demo video for better dynamic visual effects.}}   

\subsection{Quality of Fluid Video Synthesis}\label{exp-slr}

\noindent\textbf{Implementation Details.} To learn SLR, we train our multi-layered network on a series of videos containing a large region of fluid movement and still background \cite{holynski2021animating}. The dataset has only $10\%$ of its training data containing the transparent fluid pattern. To balance the data, instead of training in the whole dataset, we collect these 10\% training data as a transparent fluid subset and sample another 90\% non-transparent fluid subset. For training strategy, we train each component separately and finetune jointly. Mask loss is applied and gradually decays to zero during alpha training. For training of motion refinement network, we use the same settings in \cite{mahapatra2021controllable}. All networks are trained at a resolution of $256\times256$.

\begin{table*}[!htbp]
\resizebox{\textwidth}{!}{
\begin{tabular}{cc|ccc|ccc}
\hline
\multicolumn{1}{c|}{\multirow{2}{*}{Dataset}}                                                              & \multirow{2}{*}{Methods}    & \multicolumn{3}{c|}{All Region}                                                                                  & \multicolumn{3}{c}{Fluid Region}                                                                                \\
\multicolumn{1}{c|}{}                                                                                      &                             & \multicolumn{1}{l}{LPIPS$\downarrow$} & \multicolumn{1}{l}{PSNR$\uparrow$} & \multicolumn{1}{l|}{SSIM$\uparrow$} & \multicolumn{1}{l}{LPIPS$\downarrow$} & \multicolumn{1}{l}{PSNR$\uparrow$} & \multicolumn{1}{l}{SSIM$\uparrow$} \\ \hline
\multicolumn{1}{c|}{\multirow{3}{*}{\begin{tabular}[c]{@{}c@{}}Holynski\\ Common \\ Validation Set\end{tabular}}} & Reproduced Holynski         & 0.0798                                & 25.03                             & 0.7787                               & 0.0657                               & 25.88                             & 0.8007                              \\
\multicolumn{1}{c|}{}                                                                                      & Modified Holynski(Baseline) & \textbf{0.0793}                                & 24.75    & 0.7758     & \textbf{0.0656}      & 25.72    & 0.8000    \\
\multicolumn{1}{c|}{}                                                                                      & Ours                        & 0.0834                                & \textbf{25.14}                              & \textbf{0.7795}                               & 0.0657                                & \textbf{26.10}                              & \textbf{0.8030}                              \\ \hline

\multicolumn{1}{c|}{\multirow{3}{*}{\begin{tabular}[c]{@{}c@{}}Our\\ CLAW\\ Testset\end{tabular}}}    & Reproduced Holynski         & 0.2067                                & 20.26                              & 0.5955                               & 0.2029                                & 20.36                              & 0.5961                              \\
\multicolumn{1}{c|}{}                                                                                      & Modified Holynski(Baseline) & 0.2078                                & 19.97                              & 0.5923                               & 0.2041                                & 20.10                              & 0.5934                              \\
\multicolumn{1}{c|}{}                                                                                      & Ours                        & \textbf{0.2040}      & \textbf{20.79}    & \textbf{0.6080}     & \textbf{0.1975}      & \textbf{20.80}    & \textbf{0.6077}    \\ \hline

\end{tabular}}
\label{table1}
        \caption{\textbf{Quantitative comparison.} (a) Quantitative evaluation on Holynski's common validation set \cite{holynski2021animating}, evaluating first to 60-th frames."Fluid Region" means the static region is replaced by input image during the metric calculation to reach a higher quality result in the static background region.  (b) Quantitative evaluation of our CLAW test set.  Before evaluation, We output $768
        \times 768$ resolutions of images and resize to $640\times320$ (size of ground truth image) for our CLAW test set and resize to half size of ground truth image (around $640\times 360$) for Holynski's validation set \cite{holynski2021animating}. All the baselines are under the same ground-truth motion.}
        \label{fig:Quanti-Comparison}
\end{table*}

\noindent\textbf{Datasets and Evaluation Metrics.} For evaluation, in addition to validation set in \cite{holynski2021animating}, we collect extra $126$ scenes to form a new test set, which we called \textbf{C}omplex \textbf{L}iquid \textbf{A}nimation in-the-\textbf{W}ild~(CLAW) test set, for evaluating the performance on both dense fluids (such as rivers and oceans) and sparse or transparent fluids (such as streams or splashes). Note that, in this subsection, the motion for all the methods to be compared is guided by the optical flow of the whole video to distinguish the performance of the synthesizing techniques. For quantitative results, we use first the $60$-th frame of each sequence and compare the metric of all the frames in the middle. We use PSNR to show the whole average error, SSIM to show the errors in the region with plenty of textures and LPIPS (Alexnet version), a perceptual based loss, to show the visual quality difference. For more details, please refer to the Appendix. 

\noindent\textbf{Quantitative Comparisons.}~Table~\ref{fig:Quanti-Comparison} shows the quantitative results between our method and state-of-the-art method \cite{holynski2021animating}, which is a single layer learning-based system, a typical method that animates the scene in a global manner. We re-implement \cite{holynski2021animating} in Pytorch ({\textit{Reproduced Holynski}}), since the source code is not public available.
For other comparison baselines, {\textit{Modified Holynski}} is built upon {\textit{Reproduced Holynski}} but with the modification of replacing convolution to partial convolution in the fluid decoder.
The third model (\textit{Ours}) is our SLR model{\footnote{For \textit{Ours}, we use the (\textit{Ours (stage 1)} in Table~\ref{exp-tab2}) as surface fluid model initialization. Since our training contains three stages, and we train $100$ epochs at the first stage to get the model (\textit{Ours (stage 1)}), then  decay learning rate and fine-tune $50$ epochs for fluid at the last stage in practice, we also apply the same training strategy for the first two baselines to ensure fair comparisons. }}.
Three observations can be included from Table~\ref{fig:Quanti-Comparison}: $(1)$  PSNR and SSIM metrics of "Modified Holynski" drop a little compared to the "Reproduced Holynski" in both statistics on all- and fluid region. We infer that, although partial convolution could better capture context than conventional one to impaint reasonable texture in vacated regions in theory, simply replacing partial convolution into single-layer framework cannot \textit{essentially} improve the image quality due to the wrong context raised from improper warping of both background and fluid. While, with our SLR, the partial convolution can strengthen its advantage on context capturing for more proper fluid context textures  provided in surface fluid layer. $(2)$ The LPIPS of \textit{Ours} model drops a little compared to the \textit{Modified Holynski} model in the "All Region" of Holynski common validation set but has nearly the same LPIPS in the "Fluid Region". For other two metrics, our model leads to slightly improvements. Since Holynski common validation set contains a lot of static or opaque regions, such phenomenon tells us that the  two-layer decomposition design in \textit{Ours} model does not harm the reconstruction ability on final images. 
$(3)$ When evaluating only in the fluid region, we can see the model  \textit{Ours} reaches a comparable result with the "Modified Holynski" model in Holynski common validation set~\cite{holynski2021animating} but improves significantly in our CLAW test set. The major reason behind this observation is that when it comes to complex scenes as included in our CLAW test set, our proposed two-layer representation suppresses the influence between surrounding and surface fluids. Thus we can achieve better results than single-layer baselines that regard all elements in the scene as an entirety during warping.  
\begin{table*}[]
\small
\resizebox{\textwidth}{!}{

\begin{tabular}{cc|ccc|ccc}
\hline
\multicolumn{1}{c|}{\multirow{2}{*}{Dataset}}                                                              & \multirow{2}{*}{Methods}    & \multicolumn{3}{c|}{All Region}                                                                                  & \multicolumn{3}{c}{Fluid Region}                                                                                \\
\multicolumn{1}{c|}{}                                                                                      &                             & \multicolumn{1}{l}{LPIPS$\downarrow$} & \multicolumn{1}{l}{PSNR$\uparrow$} & \multicolumn{1}{l|}{SSIM$\uparrow$} & \multicolumn{1}{l}{LPIPS$\downarrow$} & \multicolumn{1}{l}{PSNR$\uparrow$} & \multicolumn{1}{l}{SSIM$\uparrow$} \\ \hline
\multicolumn{1}{c|}{\multirow{3}{*}{\begin{tabular}[c]{@{}c@{}}Holynski\\ Common \\ Validation Set\end{tabular}}} & Ours(Stage 1)               & \textbf{0.0782}                                & 25.11    & 0.7772     & \textbf{0.0650}      & 25.96    & 0.8026    \\
\multicolumn{1}{c|}{}                                                                                      & Ours(Stage 2) & 0.0929                                & 25.03    & 0.7612     & 0.0718      & \textbf{26.59}    & \textbf{0.8144}    \\
\multicolumn{1}{c|}{}                                                                                      & Ours                        & 0.0834                                & \textbf{25.14}                              & \textbf{0.7795}                               & 0.0657                                & 26.10                              & 0.8030                              \\ \hline

\multicolumn{1}{c|}{\multirow{3}{*}{\begin{tabular}[c]{@{}c@{}}Our\\ CLAW\\ Testset\end{tabular}}}    & Ours(Stage 1) & 0.2143                                & 20.28                              & 0.5926                               & 0.2100                                & 20.37                              & 0.5933                              \\
\multicolumn{1}{c|}{}                                                                                      & Ours(Stage 2) & 0.2411                                & \textbf{21.09}                              & 0.5674                              & 0.2294                                & \textbf{21.06}                              & 0.5738                              \\
\multicolumn{1}{c|}{}                                                                                      & Ours                        & \textbf{0.2040}      & 20.79    & \textbf{0.6080}     & \textbf{0.1975}      & 20.80    & \textbf{0.6077}    \\ \hline
\end{tabular}}
\label{table_ablation}
        \caption{\textbf{Quantitative Abalation.} (a) Quantitative evaluation on Holynski's common validation set~\cite{holynski2021animating}. (b) Quantitative evaluation of our CLAW test set. Unless otherwise specified, all settings are the same as Table ~\ref{fig:Quanti-Comparison}.}\label{exp-tab2}
\end{table*}

\noindent\textbf{Quantitative Ablation.} Table~\ref{exp-tab2} shows the quantitative ablation among each traning stage of our SLR model. Network input resolution is $768 \times 768$ to get higher resolution results during inference. For evaluation, We resize these $768 \times 768$ images to the size of the ground truth image(or half of it in Holynski common validation set). The \textit{Ours (stage $1$)} is trained under the same setting as the model (\textit{Modified Holynski(Baseline)} in Table~\ref{fig:Quanti-Comparison} but with lesser epochs. The prediction of \textit{Ours (stage $2$)} model is a uniform blending of the surface fluid image from the stage $1$ model and background image from background extractor trained in stage $2$, telling us a naive combination will lead to blurry fluid synthesis in the non-transparent fluid region of the animating videos. The \textit{Ours} is the final SLR model, which has comparable results with the best ones of {Ours (stage $2$)} or {Ours (stage $1$)}  model in Holynski common validation set , and improves significantly in our CLAW testset.

\begin{figure*}[!t]
	\centering
	\includegraphics[width=\textwidth]{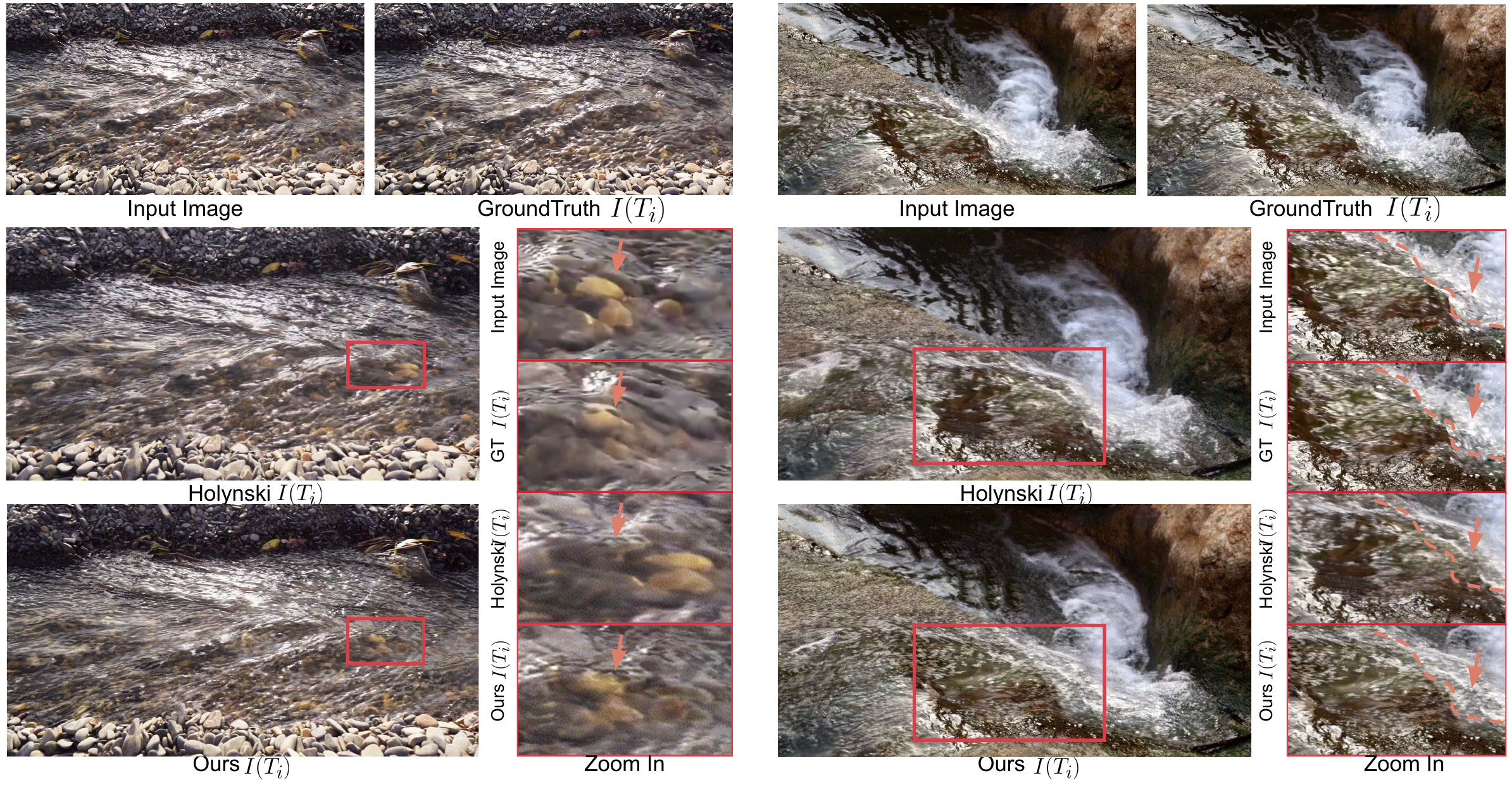}
	\caption{\textbf{Qualitative comparison with previous SOTA method}. Two complex transparent fluid scenes are visualized. Left sample: rock under fluid is moving w/o our SLR method. Right sample:  {\color{pink} dotted line}  points out the boundary between the transparent solid region and non-transparent fluid region. Two methods show comparable results in non-transparent region. While, for transparent one, \cite{holynski2021animating}'s method leads the texture of intertidal zone moves with the fluid flow . In contrast, our method properly keeps the background still.}
	\label{fig:ComparisonSingleLayer}
\end{figure*}

\noindent\textbf{Qualitative Comparisons.} Figure~\ref{fig:ComparisonSingleLayer} shows more comparisons between single layer and our two-layer models. Our model successfully decouples rock textures and fluid textures above the rock so that when the liquid moves after a period, the texture beneath the liquid appropriately stays still in our representation, rather than moving with fluids in the single-layer model \cite{holynski2021animating}.

\begin{table}[h]
	\centering
	\begin{tabular}{l|ccc}
	\hline
	&Single layer & Ours & Neither \\ \hline
	E-O-N 	& 20.5\% 	& 52.3\% & 27.2\%     \\ \hline
	\end{tabular}
	\vspace{1mm}
	\caption{\textbf{User study.} The subjective scores are voted on all samples of our CLAW testset, which includes $122$ videos with various real-world fluids scenarios(\textit{e.g.,} river, stream and waterfall) and complex context surrounding (\textit{e.g.,} semi-/transparent liquid, collisions and thin structures).}
	\label{table2}
\end{table}

\noindent\textbf{User Study.} We conduct a user study to compare visual effects between single layer baseline~(\textit{Modified Holynski} in Table~\ref{table1}) and our two layers method(\textit{Ours} in Table~\ref{table1}) on the full set of {\textit{CLAW testset}}. For each scene we animate $60$ frames and output a two second video. 
Animated videos are shown side by side without providing ground truth videos as reference to guide the participator focusing on the reasonability and photo-realistic. We ask several participants to select which one is better or neither is better, mainly according to three dimensions (photo-realistic, stereoscopic and high-fidelity).

As shown in Table~\ref{table2}, the participators can make a distinction between the two methods over most samples, leaving $27.2\%$ samples difficult to tell. Our method achieves better scores than the single-layer method~\cite{holynski2021animating} with $52.3\%$ sample voting. This indicates the effectiveness of the proposed SLR in other aspects.

\begin{figure*}[htbp]
	\centering
	\includegraphics[width=\textwidth]{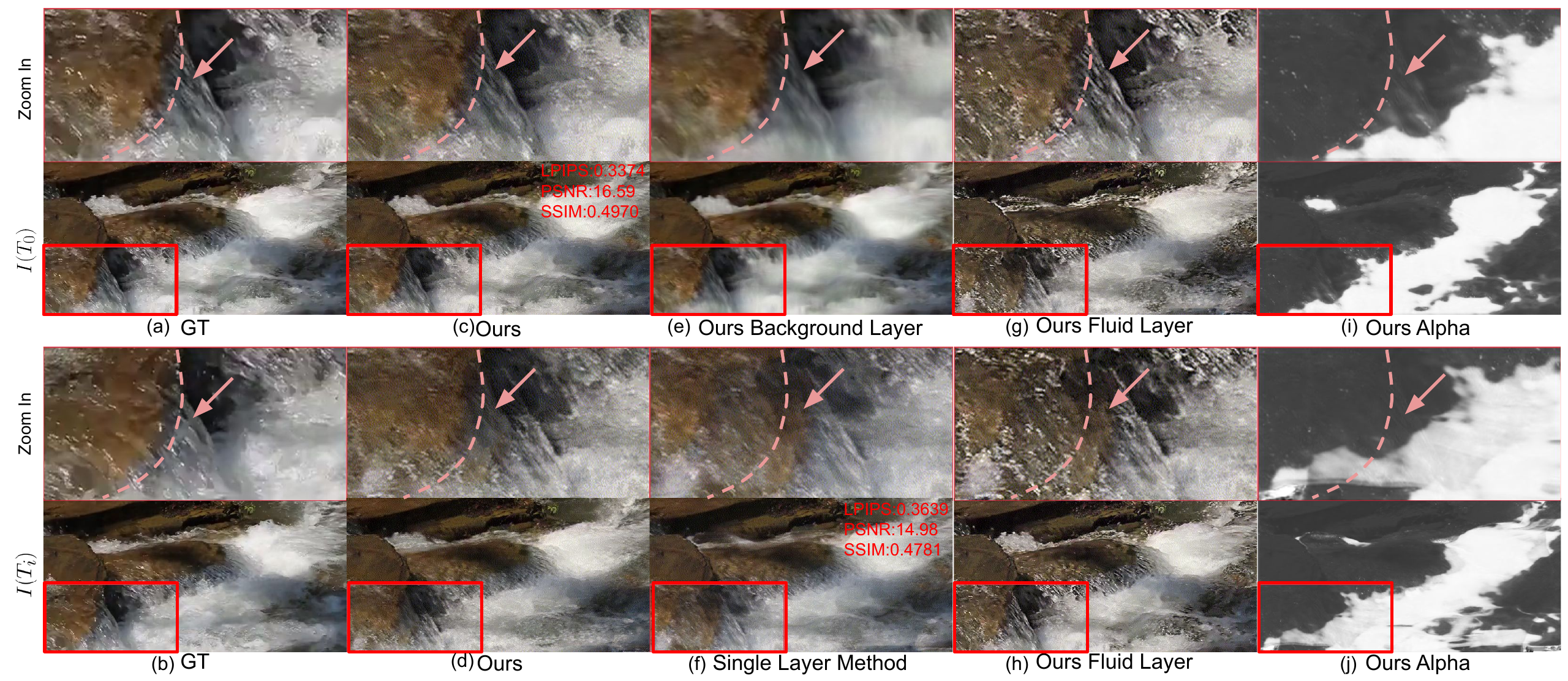}
	\caption{{\textbf{Visualization of decomposition results.}} The figure shows all components of our final prediction, explaining how our SLR leads to realistic fluid animation even under a complex transparent scene, with only moving fluid textures above rocks. (a-b)Ground-truth frames $I(T_0)$ and $I(T_i)$. (c-d) Ours final prediction for $(T_i)$. (e) Background layer prediction. (f) Single-layer method~\cite{holynski2021animating} prediction.  (i-j) Composited alpha prediction, alpha is $1$ at surface fluid textures only region and is small at rock textures region.  {\color{pink} Pink arrows and dotted line}  point out the boundary of the transparent solid region and non-transparent fluid region. We can see the rock is moving in single surface fluid layer(f) and (h) but keep static in Ours(d). } 
	\label{fig:AllResults}
\end{figure*}

\begin{figure}[htbp]
	\centering
	\includegraphics[width=\linewidth]{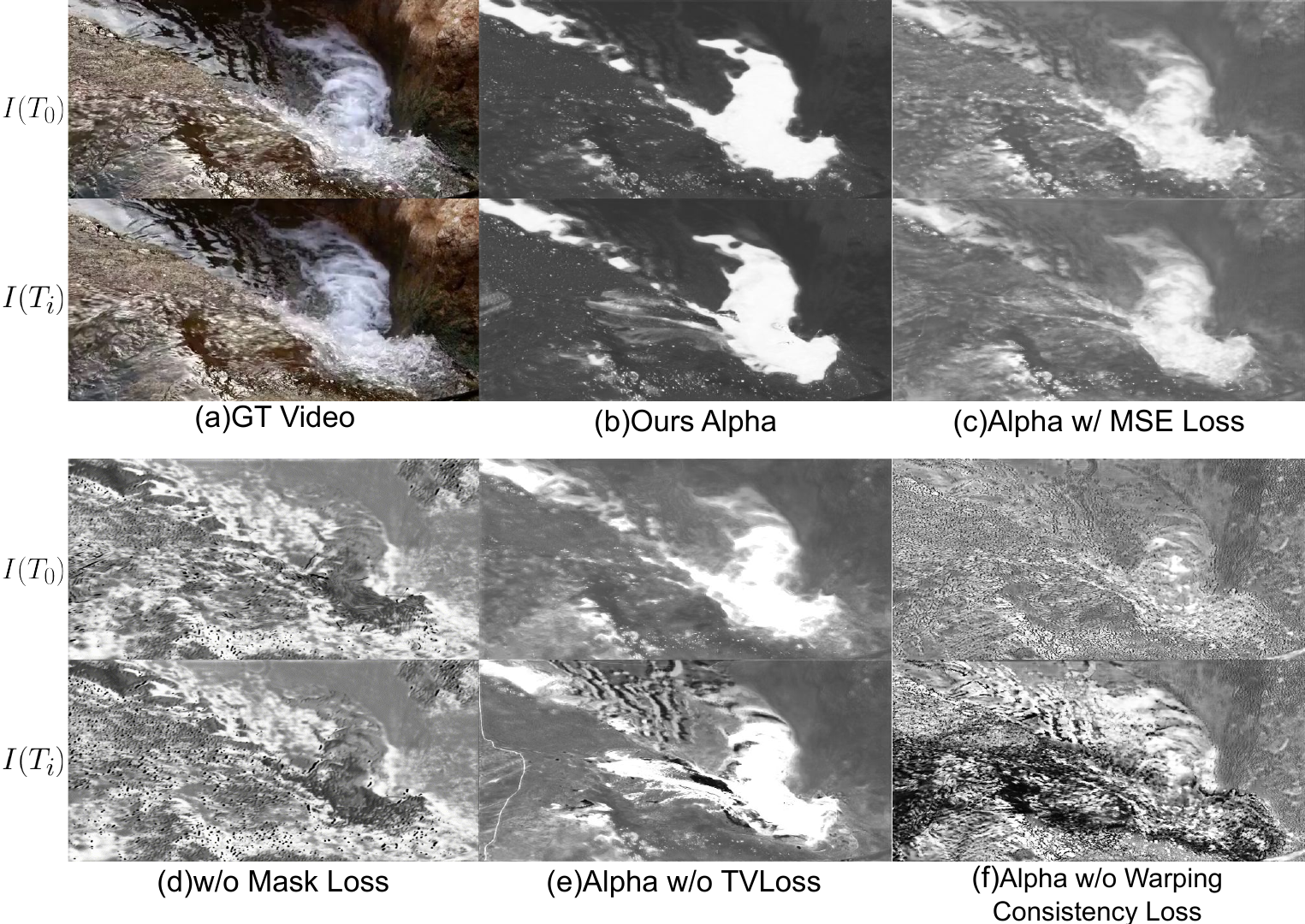}
	\caption{{\textbf{Influence of Different Losses to $\alpha$ Learning.}}~From (b,c) can tell that L1 loss gives a more accurate alpha than MSE loss in our experiments. (d)Training alpha with reconstruction loss cannot predict an accurate alpha in the spray and splash region. (e)Alpha total variation loss alleviates alpha noise. (f)Training without alpha warping consistency loss cannot guarantee a temporally consistent output.
}
	\label{AlphaLearning}
\end{figure}

\noindent\textbf{Decomposition Results.} To help understand the  network estimation for proposed representation in a more straightforward manner, we visualize a complex fluid case with each component predicated from our model, as shown in Figure~\ref{fig:AllResults}. In the almost transparent fluid region (left side of dotted line in the figure), the surface fluid layer (g-h) carries more high-frequency fluid textures above rock compared with single-layer method (f), and the background layer removes most surface fluid textures compared with input image(a). alpha(i-j) is small but not zero in the region, associated with the static background and moving foreground, combines to meaningful transparency. 
 As shown in the pink arrow, the movement of rock under fluid is suppressed in our final prediction(d), while single-layer method (f)~\cite{holynski2021animating} animates improperly with both rock textures and fluid textures moving.

\noindent\textbf{Influence of Different Losses to $\alpha$ Learning.} As shown in Figure~\ref{AlphaLearning}, for the learning of alpha factors, our labelled mask indicating the transparent region helps the alpha factors to converge to a semantic result. An absolute error(L1) on the output alpha with the labelled mask may extract sharp transparent factors than the squared mean (MSE). We can also find that total variance restriction strengthens the smoothness of the alpha channel to fill holes caused by warping, and warping consistency makes the alpha learning more aligned with input image textures, lessening the ghosting effects and blur in the final output.

\subsection{Different Motion Prediction} \label{exp-sfs}

\begin{figure*}[!htb]
  \centering
  \includegraphics[width=\linewidth]{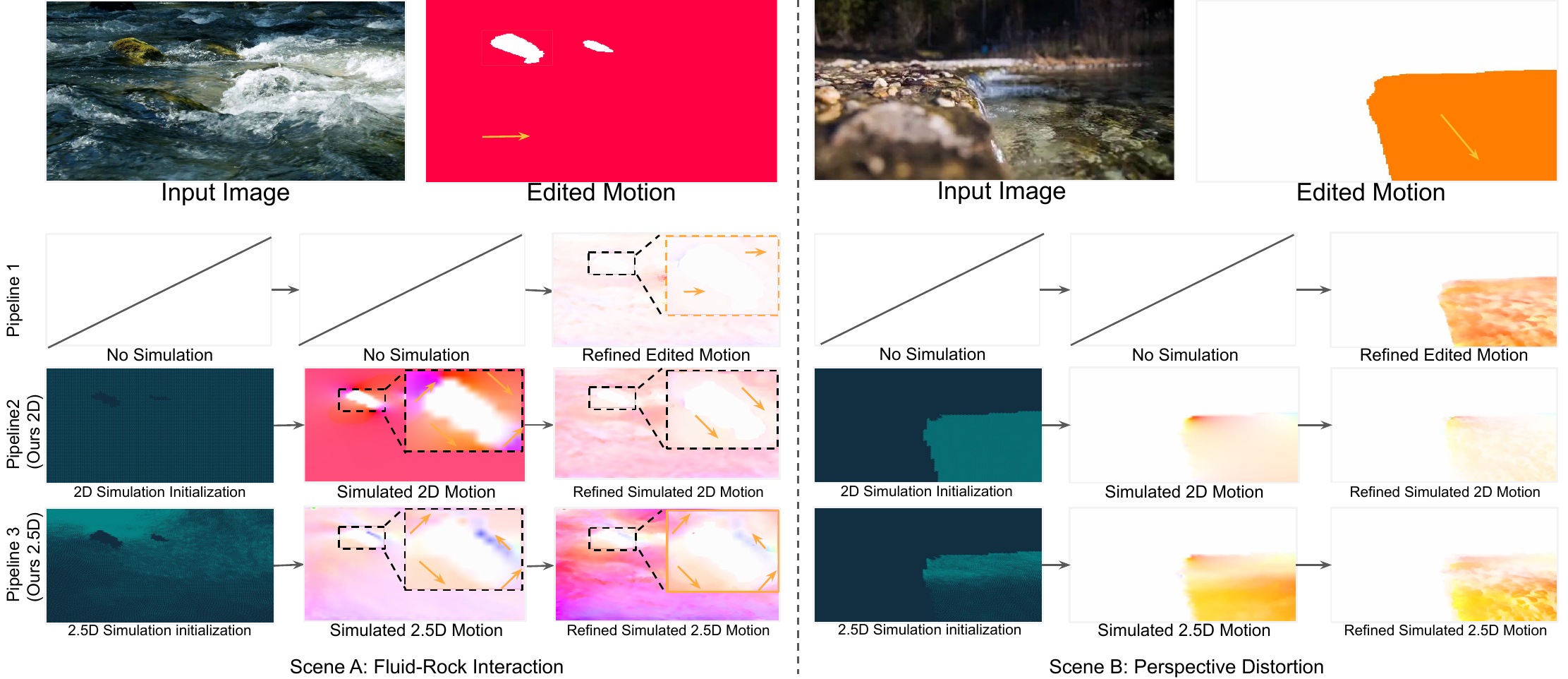}
  \caption{\textbf{Different simulation methods.} Orange arrows point out the motion direction. All simulation methods use 75\% week incompression simulation to get a better visual effect in this scene. (row 1)Input image, sparse motion generated by sparse user hint required by each method. (row 2-4)ablation study of Mahapatra's method~\cite{mahapatra2021controllable}, 2D simulation and 2.5D SFS.
}
\label{fig:SimulationAblation}
\end{figure*}

Figure~\ref{fig:SimulationAblation} shows the different motion prediction pipelines. {\textbf{Pipeline $1$:}} An interactive sparse labelling of the motion and then extending all the velocities to the whole moving region, which results in a constant flow that reminds large gap with realistic motion. We can see that the motion appears reasonable global with a network refinement. However, for local regions around static rocks, the flow of liquid is improperly stiff and has no interactions with the rock. While in real-world scenes, there should have vortex and sprays caused by collisions. {\textbf{Pipeline $2$:}} Built upon pipeline $1$, while before the motion refinement stage, we additionally initialize grids with fluid, rock or air masks as well as placing particles around each grid, then advocate particles by edited motion for once. Later, an NS equation with the incompressible fluid assumption is solved on the 2d grid to obtain the motion of the next frame, which we call "Simulated 2D Motion" in the figure. Considering pipeline $2$ is simulating on a 2D image plane, a refinement step (using a smooth motion translator) is applied to compensate for the motion offset at height and fine details. {\textbf{Pipeline $3$:}} instead of simulating on a 2D image plane, the N-S equation is solved in $2.5D$. Specifically, we regress the depth with a pretrained monocular depth estimator and consequently build a 3D mesh over the fluid. Then, we solve the N-S equation on the mesh surface. With the help of simulation, we can see in the left scene of the figure that the effect of turbulence around the rock can be animated, and the trend maintains the refined output. From the right scene in the figure, we can observe that the absolute value of the velocity is unnatural in Pipeline $2$, which regards all particles on an identical plane. However, since the liquid surface leans in the picture, there should be a foreshortening effect on the projection. 
In contrast, pipeline$3$ samples particles on the surface and such simulation enable the near-plane motion to carry more speed.

\section{Byproducts}\label{sec-bp}

\begin{figure}[!htbp]
	\centering
	\includegraphics[width=\linewidth]{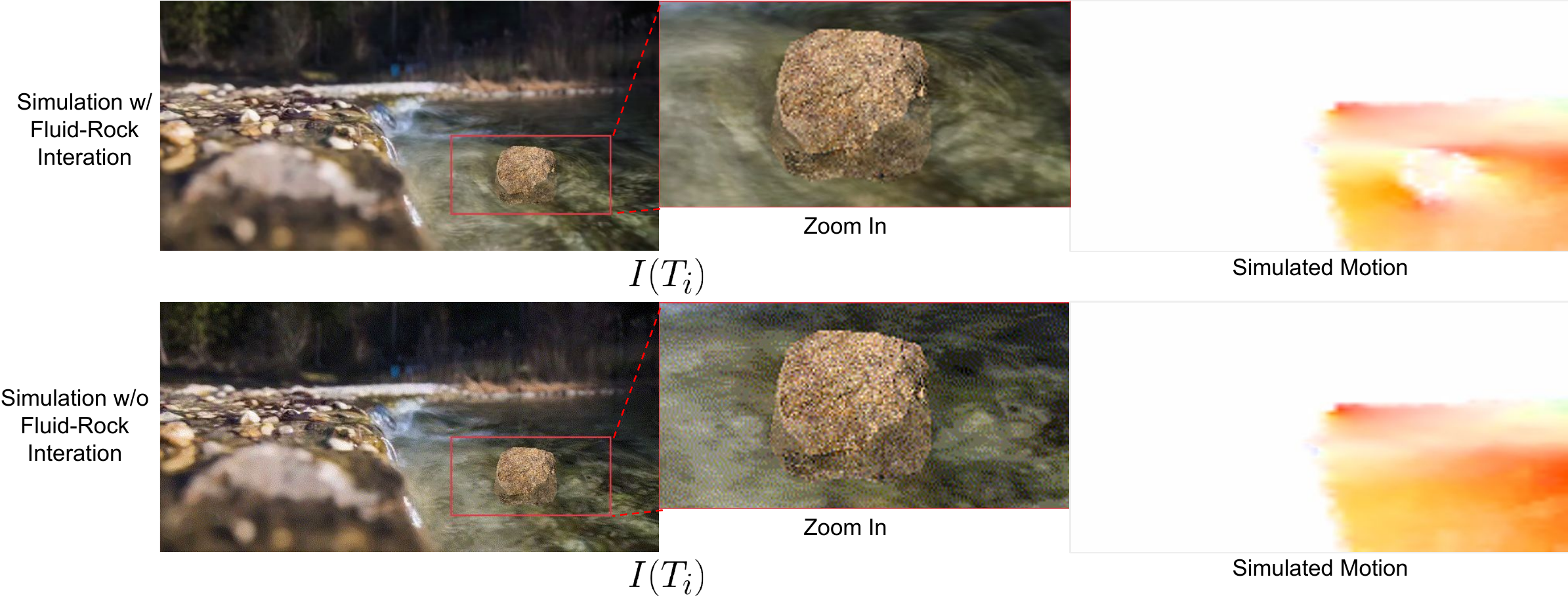}
	\caption{{\textbf{Byproduct: interactive editing.}} Here we present rock editing with fluid-rock interaction simulation. (Down)Initial all moving region as 'Fluid' statue. (Up) Initial 'Solid' statue instead of 'Fluid' statue at overlap area of mesh and the new object. 
}
\label{fig:SimulationFluidRockInteration}
\end{figure}

As our system naturally disentangle the scene and marries the advantages of both physic-and learning-based simulation, we can generate multiple AR effects on a given fluid image. Here, we list some examples.

\noindent\textbf{Interactive Editing.} We can augment the fluid scene while keeping realistic fluid animation. For example, we can place an imaginary stone in a river that is originally slack water, and the flow could be updated to the new one with the effect of a vortex around the stone, as can be viewed in {\textit{Simulation w/ Fluid-Rock Interaction}} of Figure~\ref{fig:SimulationFluidRockInteration}. To achieve this, we only need to calculate the interface of inserted stone mesh and fluid surface mesh. Then, we classify the part of the stone above and beneath the river surface according to the depth. Later, the motion of the fluid is updated with boundary conditions during simulation. Please refer to Appendix~\ref{ref-formula} for more implementation details. 

\begin{figure*}[!htbp]
	\centering
	\includegraphics[width=\linewidth]{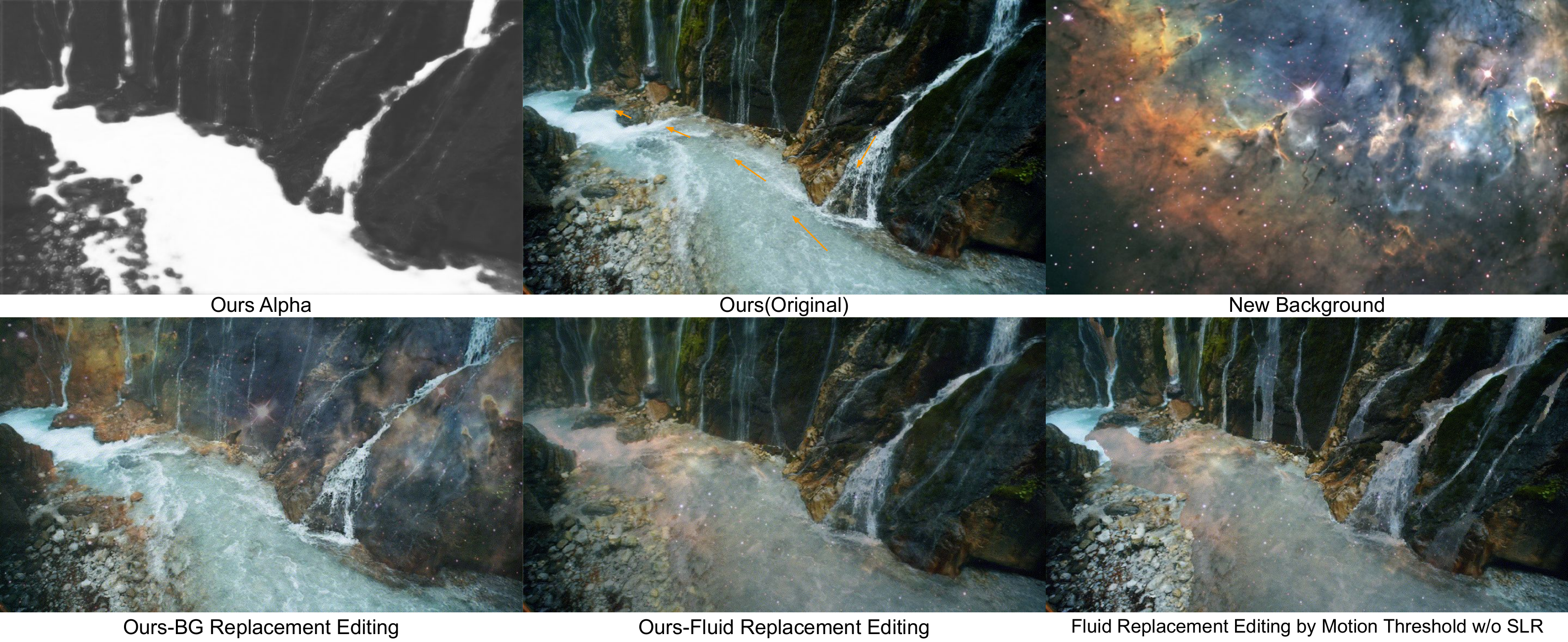}
	\caption{Layer Replacement Editing. The first row show our SLR output and new background. The second row shows results of our background layer replacement effects( in this scene, background layer is replaced by 0.25$\times$original background layer + 0.75$ \times$new space background layer), fluid layer replacement effects( in this scene, fluid layer is replaced by 0.5$\times$original fluid layer + 0.5$\times$new space background layer), and replacement effects using motion instead of alpha(in this case, we replace moving region by 0.5$\times$original Holynski's output + 0.5$\times$new space background layer). {\color{orange}Arrows} specify the motion direction and speed. 
}
\label{fig:LayerReplacement}
\vspace{-4mm}
\end{figure*}
\noindent\textbf{Layer Replacement Editing.} We can change texture attributes of both surface fluid and background by simply replacing either fluid layer or background layer with another reference texture. For example, 
as shown in Figure \ref{fig:LayerReplacement}, the fluid layer can be replaced by/regrouped with a starry sky video with an endless loop to form dreamlike fluid effects that might remind the user of a Chinese classical poetry---\textit{as if the Silver River fell from azure sky}. For a downstream task like such texture attribute replacement, accurately segmentation of all fluid regions sometimes is necessary to reach plausible visual effects. However, it is hard to be accomplished by brute force segmentation with either motion prediction (\textit{Fluid Replacement Editing by Motion Threshold w/o SLR} in the figure) or manually labelling. Since real-world fluid scenes contain complex context relations
such as semi-transparency, different motion behaviour among fluid regions in one scene and so on. On the contrary, since our proposed surface-based layer representation (SLR) naturally disentangles the scene into surface fluid and background, we can achieve such editing easily.

\section{Conclusion}
We propose a fluid simulation pipeline that decomposites a still image into surface liquids and backgrounds, and synthesizes the animating video with fluid motion. Compared to traditional warping pipeline, the decomposition of liquid and objects can better simulate the transparent fluids as well as other fluids scenes with complex surrounding, and generate a semantic decomposition factor. In addition, our simulation method provides a wider range of editing abilities for image and video application. Although our method works in a semi-supervised way, with little supervision from ideal decomposition or graphic based rendering, the actual outputs may have some artifacts in some special view of cameras. We expect more research on how to naturally acquire the fluid-background splitting, and how to apply more precise motion prediction.

\noindent
\textbf{Acknowledgements.} This work is supported in part by Centre for Perceptual and Interactive Intelligence Limited, in part by the General Research Fund through the Research Grants Council of Hong Kong under Grants (Nos. 14204021, 14207319, 14203118, 14208619), in part by Research Impact Fund Grant No. R5001-18, in part by CUHK Strategic Fund.

{\small
\bibliographystyle{ieee_fullname}
\bibliography{egbib}
}

\newpage

\begin{appendices}

\section*{Appendix}
\section{Dataset}
\subsection{Holynski Dataset}

\begin{figure*}[!h]
	\centering
	\includegraphics[width=\textwidth]{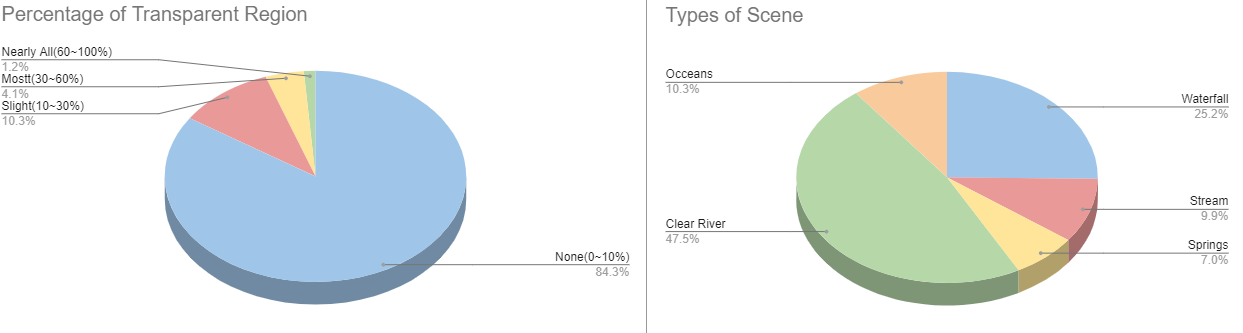}
	\caption{Main statistics of Holynski's training set.}
	\label{table:Hdataset}
\end{figure*}

\begin{figure*}[!h]
	\centering
	\includegraphics[width=\textwidth]{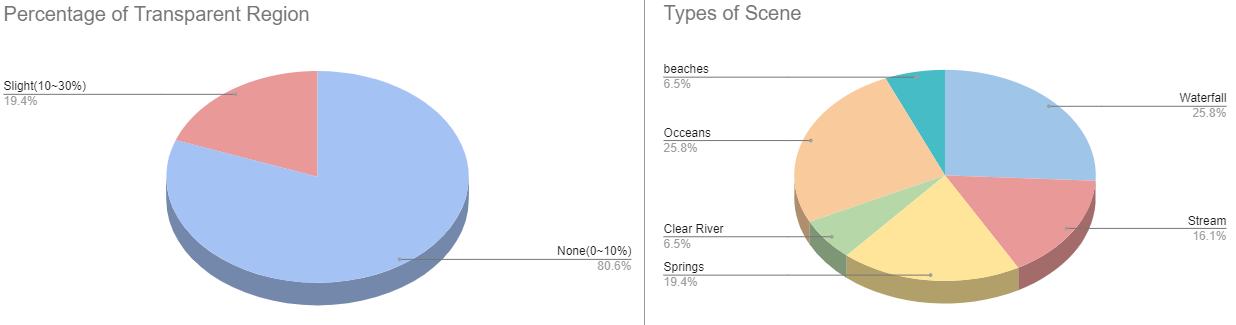}
	\caption{Main statistics of Holynski's validation set.}
	\label{table:Hvaldataset}
\end{figure*}

The dataset proposed by \cite{holynski2021animating} includes a variety of fluids under natural scenes, such as waterfalls, oceans, and fogs. 
We regard all the training samples ($949$ scenes, $5$ short video clips for each scene on average) as the training set and perform testing on the validation set{\footnote{since \cite{holynski2021animating} does not release ground-truth videos for their test set, we use the validation set for evaluation.}} ($31$ scenes). The main statistics of this dataset can be seen in Figure~\ref{table:Hdataset} and Figure~\ref{table:Hvaldataset} . Although diverse real-world fluids are covered in this dataset, most of the scenes are opaque.

\begin{figure*}[!t]
	\centering
	\includegraphics[width=\textwidth]{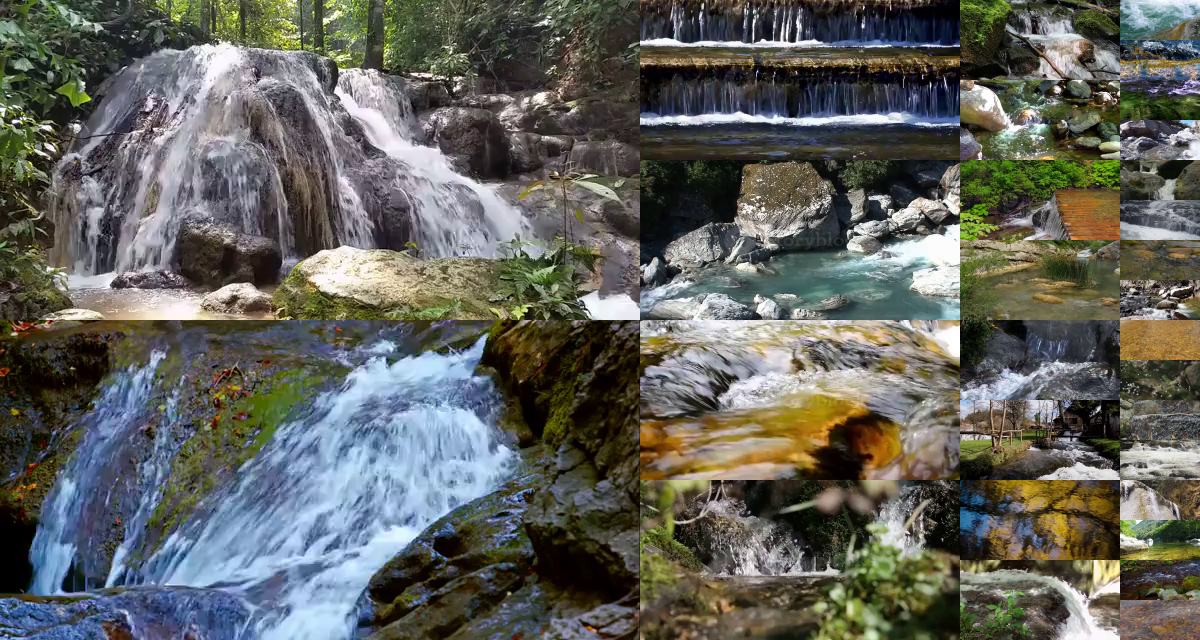}
	\caption{\textbf{Samples in our proposed CLAW test set.}}
	\label{fig:dataset}
\end{figure*}

\begin{figure*}[!h]
	\centering
	\includegraphics[width=\textwidth]{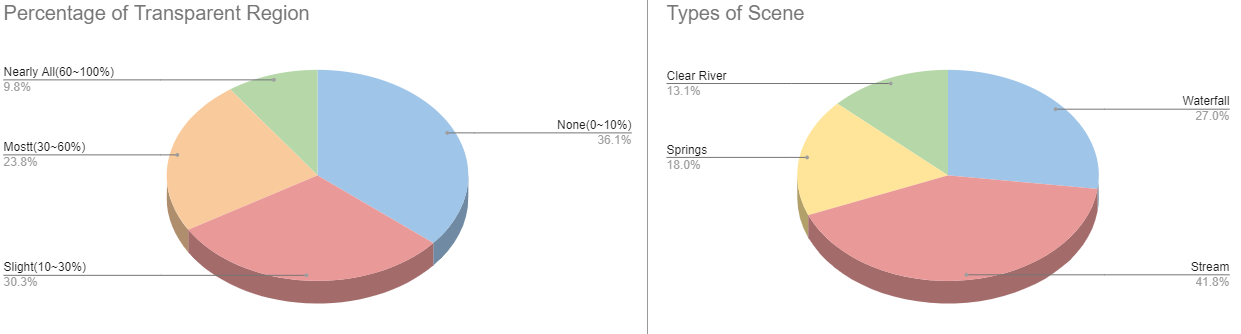}
	\caption{Main statistics of proposed CLAW test set.}
	\label{table:dataset}
\end{figure*}

\subsection{Our Proposed CLAW Test Set}
To further quantitatively evaluate our method on more complex scenes(\textit{e.g.,} semi-/full transparent fluids) and facilitate future research of fluid animation, we collect a new test set named Complex Liquid Animation in-the-wild (CLAW) from StoryBlocks\footnote{https://www.storyblocks.com/} with key words \textit{fluid, waterfall, river etc.} and filter them manually to exclude unrelated videos. Videos from $122$ scenes are finally provided. We follow two rules to select these videos. First, there must have transparent/semi-transparent fluid regions in the scene. Second, the motion of transparent fluid should be predicted reasonably by a pre-trained optical flow estimator, such that we can provide pseudo-ground-truth motion fields for network learning. We use flownet2~\cite{ilg2017flownet} in practice. The main statics of our dataset are shown in Figure \ref{table:dataset}. Compared to the Holynski validation set, more challenging context relations with fluids are provided in the CLAW dataset, and the type of fluids and transparent regions are more balanced in CLAW. Some image samples of our test set are presented in Figure~\ref{fig:dataset} .

\section{Implementation Details}

\subsection{Generating Labels for Alpha Values}
As mentioned in the main paper, a few labels of transparency values are expected to initialize the learning of $\alpha$. Thus, we re-annotate around $600$ masks in the Holynski training set(note: we pick one image frame per scene). The definition of these masks is that the pixels contain solids that are overlapped with fluid regions. Then, for these masked regions, we set labels as $\alpha_{bg}^{gt}$ = $0.25$.
 For fluid regions with motion value greater than $0.1 \times $ mean of motion speed , we set the labels as $\alpha_{fluid}^{gt}$ = $1$. The weight of this supervision gradually decays to zero during alpha training.

\begin{figure*}[!t]
	\centering
	\includegraphics[width=\textwidth]{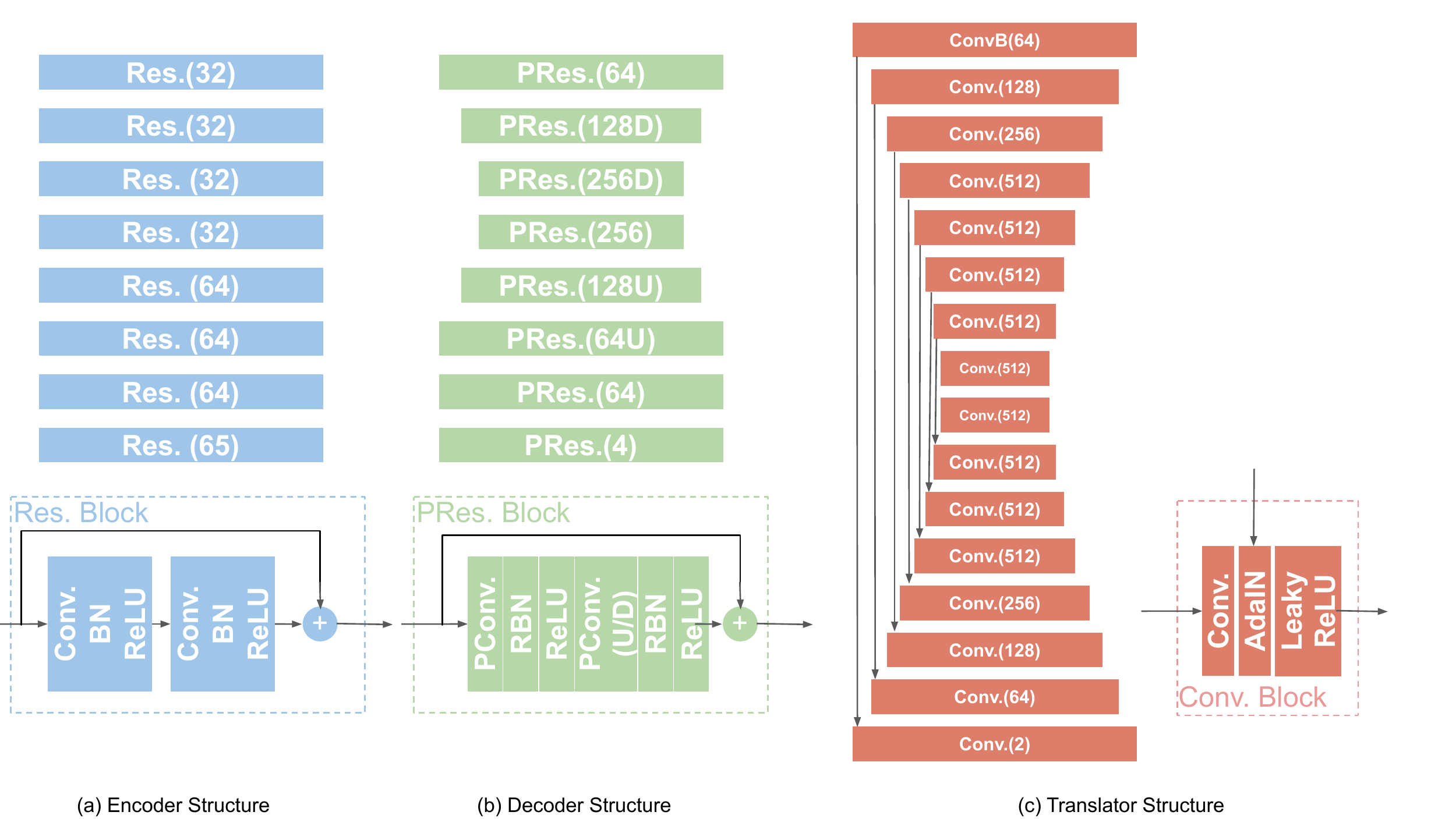}
	\caption{\textbf{Architectures.} (a) The structure of the encoder network, with 8 res-blocks. Each consists of a pair of convolutions, batch-normalization and relu layer. No upsampl or downsample is used here. The number in the bracket shows the output channels (b) The structure of the decoder network, with 8 partial res blocks. Each consists of a pair of convolutions, region-based batch-normalization(normalized on regions with positive mask), and relu layers. The upsample(U) and downsample(D) are done with strided convolutions or deconvolutions on the second unit of the res-block. (c) The structure of the translator transfers an image and guided flow map to a refined flow map. Each conv-block is built with a convolution, an adaptive instance normalization, with mean and variance acquired from guided dense flow needed to be refined. The activation is set to be Leaky-Relu to smooth the network.}
	\label{fig:network}
\end{figure*}

\subsection{Networks}

In this subsection, we provide more details of the network consisting of three major components, an \textit{Encoder} that maps the images to the corresponding background image, fluid layer features, $\alpha$ channel and $Z$ channel, a \textit{Decoder} that refines the warped features to final surface fluid layer image and the warped alpha to final alpha, and a \textit{Translator} that refines the coarse velocity to the one with fine-grain.

\noindent\textbf{Architecture.} For the encoder, we use $8$ ResNet Blocks, which is the same as Synsin~\cite{Wiles_2020_CVPR}. For the decoder, we use $8$ ResNet Blocks and replace the convolution operator with partial convolution \cite{liu2018image} along with mask input. The mask is vacated region of warped features or warped alpha. Partial convolution 
helps inpaint irregular blank areas caused by warping. For motion translator, we use U-net with $16$ layers of convolution and use SPADE layer \cite{park2019SPADE}instead of batch-norm. The detailed structure can refer to Figure~\ref{fig:network}.

\noindent\textbf{Training.} As described in Section $3.1$ in the main paper, we train each part of the model separately and jointly train together afterwards. For the first and second stages, the learning-rate of the generator and discriminator is $5e-4$ and $2e-3$ , with the loss weight of the components to be L1: $1.0$, Perceptual: $10.0$ and GAN: $1.0$. For the final stage, with the previous trained network prior, the learning rate is $2.5e-4$ and $1e-3$, and a weighted L1 Loss with a weight $30.0$ is added.

For training the translator network, 
we follow the training strategy in \cite{mahapatra2021controllable}, but with some modifications. Specifically,  we concatenate the source fluid image, initial dense motion map and the moving fluid region mask to form the input tensor and feed the tensor to the translator network. The output of the network is supervised with pseudo-ground-truth motion. The loss terms contain endpoint error, with weights $10$ and GAN loss, with weights $1$ . We use a learning rate of $5e-4$ and $2e-3$  to train the network and finetune with a fluid layer with a learning rate of $2.5e-4$ and $1e-3$.

\subsection{Editing Pipeline}

As described in Section~\ref{exp-sfs} in the main paper, we can edit the fluids with imaginary objects and change the flow of the liquids. We detail the implementation process of such an editing application in this subsection.

As shown in Figure~\ref{fig:motion_pipeline} in the main paper, the fluid mask (step $2$) and monocular depth (step $4$) is generated directly from the input image, and the initial dense velocity map (step $4$) is generated with the user interactive sparse labelling (step $2$). With the mesh (step $5$) built upon the depth map and isotropic triangularization, we insert the cad model of the target object into the 3D scene at a proper position that crosses with the scene mesh(step $9$). 
Then a Z-buffer algorithm is applied to detect the occluded region of the scene mesh and the object (step $12$). For velocity on mesh, we cut the mesh with the occluded area viewing as a solid boundary. The surface-only fluid simulation is performed with updated boundary conditions on the new mesh to achieve the effect of fluid colliding with solid, and the simulated velocity is refined with the pre-trained translator (step $13$) to obtain the final motion fields. To obtain the final animated video, we need to render the edited scene. Specifically, we set the unoccluded region in front of the fluid layer to show the immersion effect (step $10$) and the occluded one of the objects into the background layer (step $11$). Then the warping is done the same as in the previous pipeline (step $14$). With the help of our two-layer model and simulated velocity, we can edit the fluid image with various effects. Please refer to our supplementary video and project page for more details.

\section{Formula for Surface-only Fluid Simulation}\label{ref-formula}

Traditional fluid simulations usually use grid-based sampler and simulate the fluids according to Euler Equation with no stickiness assumption. The overall equation is formulated as 
\begin{align}
	\frac{\partial u}{\partial t} + u\cdot\nabla u &= -\frac{1}{\rho}\nabla p + g \nonumber \\
	\nabla\cdot u &= 0
	\label{NS_sup}
\end{align}
where $u$ is the fluid velocity, $\rho$ is the density, $p$ is the pressure inside the fluid, and $g$ is the external force, where only gravity is considered here. The second equation presents the incompressibility properties.  

The PIC method \cite{zhu2005animating} can be applied to solve equation \ref{NS_sup}.
This equation can be splitted by three steps: advection, apply forces and pressure projection(incompressibility)~\cite{bridson2015fluid}. Since the key to our system is pressure projection and the other steps are not necessary, we only introduce how to solve this pressure projection equation:
\begin{align}
	\frac{\partial u}{\partial t} &= -\frac{1}{\rho}\nabla p \nonumber \\
	\nabla\cdot u &= 0 
	\label{PressureProjection}
\end{align}
This equation can also be expressed as a Poisson's equation form, which is easier to solve . We addtionally add boundary conditions to the Poisson's equation~\cite{bridson2015fluid}:
\begin{align}
	\nabla \cdot \nabla p &= \frac{\rho}{\Delta t} \nabla \cdot u \nonumber \\
	 u \cdot n &= 0 \qquad at  \; solid  \; boundaries \nonumber\\
	 p &= 0 \qquad at \; free \; surfaces
	\label{Possion's equation}
\end{align}
where $n$ is the normal of labelled solid and $\Delta t$ is time step. 

At the heart of our SFS is to solve this Poisson's equation on mesh, which will be detailed later. After solving the pressure $p$ and its gradient $\nabla p$,  the velocity satisfying incompression can be calculatad by:
\begin{align}
	u^* &=  u - \frac{\Delta t}{\rho} \nabla p
	\label{PressureProjection2}
\end{align}
where $u$ is the fluid velocity corresponding to the motion map predicted by our translator network, and $u^*$ is the fluid velocity after pressure projection. We use $u^*$ to update our motion map.  

\begin{figure*}[!t]
	\centering
	\includegraphics[width=\textwidth]{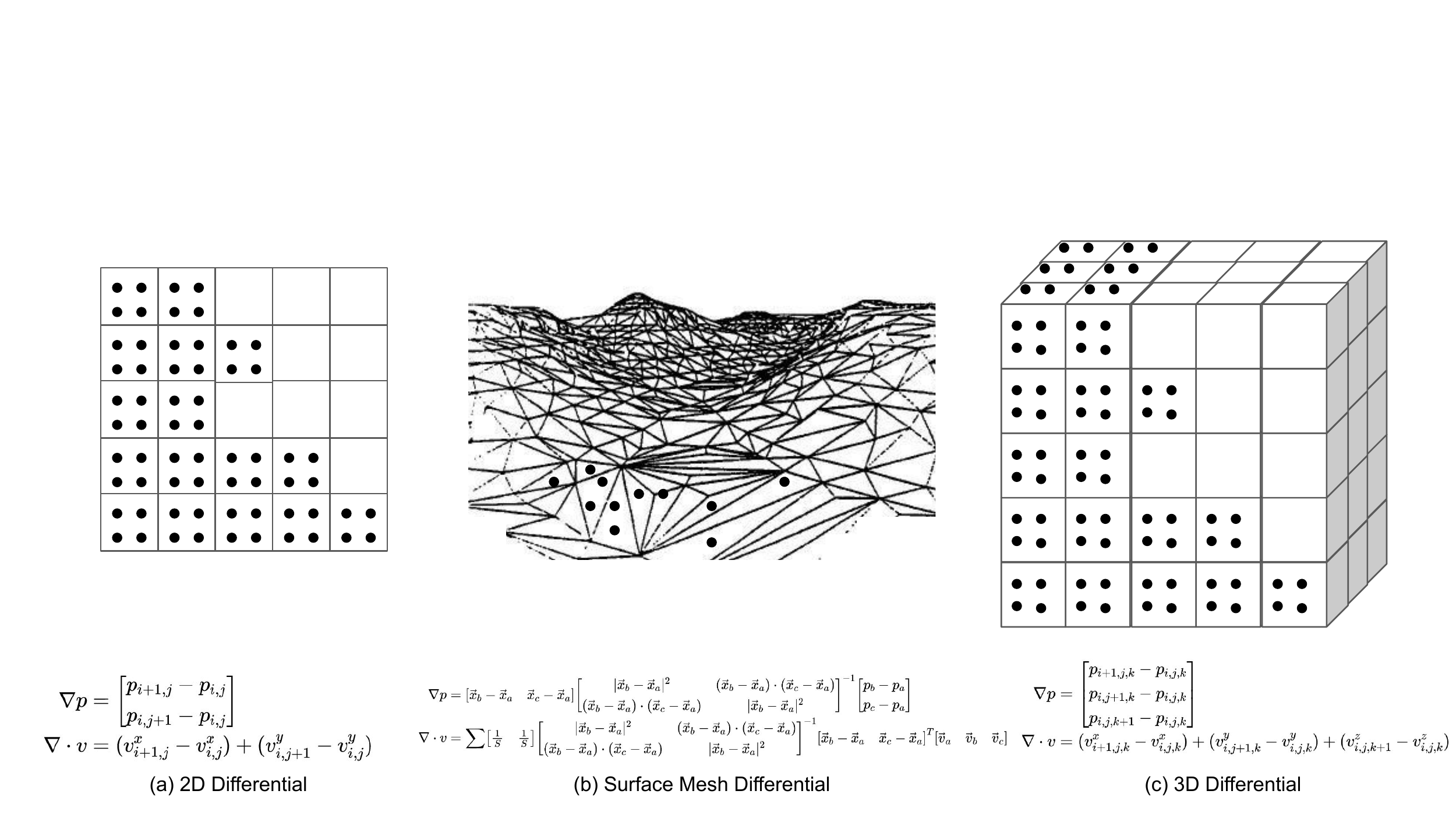}
	\vspace{-6mm}
	\caption{\textbf{A toy example of different formula settings.} Here we show the different formulas for gradient and divergence calculation in 2D/Mesh/3D. The sampled material points are regularly distributed on each unit. For 2D/3D version, the differentials are easy to be discretized as finite subtraction, while on the mesh-based surface, we need to use the vertex position to formulate a connection between the differential and the vector field. When the mesh grid is flattened on a 2D plane, we can derive the formula to be proportional.}
	\label{fig:formula}
\end{figure*}

Instead of performing previous steps on 2D or 3D grids on vertex, we perform the same procedure on a mesh layer representation, which has depth information, but only a slice of the surface is simulated.  A toy example of the differences among proposed ones with traditional 2D and 3D equations is shown in Figure~\ref{fig:formula}. 

Specifically, in our framework, the mesh is built for the whole image with a mask region of fluids using mono-depth estimation and perspective projection. The vertices are calculated with depth as 
\begin{align}
	x &= (u - cx) / fx \cdot d \nonumber\\
	y &=-(v - cy) / fy \cdot d \nonumber\\
	z &= d
\end{align}
where $x$,$y$,$z$ is the extracted vertex position, $u$,$v$ is the pixel coordinate on the image, and camera parameters are set to be $fx$,$fy$,$cx$,$cy$ as a perspective camera with a field of view(FOV) angle of 90 degrees in height.

With the given 3D surface and its projection information, we need to estimate the initial flow of the scene. A 2D flow field is generated with the help of the users' label, as described in Section 3.2 in the main paper. Assuming the projected velocity is consistent with input flow for each triangular surface, the equation is set to be
\begin{align}
	v_{\Delta abc} = \begin{bmatrix}
			\mu & \lambda
		\end{bmatrix}
		\begin{bmatrix}
			x_{b}-x_{a} & x_{c}-x_{a} \\
			y_{b}-y_{a} & y_{c}-y_{a} \\
			z_{b}-z_{a} & z_{c}-z_{a}
		\end{bmatrix}
\end{align}
where $abc$ is the vertex indices for each triangular face, $x,y,z$ is the 3D position, and $v$ is the final velocity, the direction of the velocity must be parallel to the plane defined by $\Delta abc$, therefore a linear combination of the edges. We need to differentiate the projection equation
, which is
\begin{align}
	\frac{d}{dt}\begin{bmatrix}
		u \\ v
	\end{bmatrix} &= \frac{d}{dt}\left(\begin{bmatrix}
		fx & 0 & cx \\
		0 & fy & cy
	\end{bmatrix}\begin{bmatrix}
	x/z \\ y/z 
	\end{bmatrix}\right) \nonumber \\
	&= \begin{bmatrix}
		fx & 0 \\
		0 & fy
	\end{bmatrix}\begin{bmatrix}
		1/z & 0 & -x/z^{2} \\
		0 & 1/z & -y/z^{2}
	\end{bmatrix}\begin{bmatrix}
		\frac{dx}{dt} \\
		\frac{dy}{dt} \\
		\frac{dz}{dt}
	\end{bmatrix}
\end{align}
where the velocity $v =\begin{bmatrix}\frac{dx}{dt} \\ \frac{dy}{dt} \\ \frac{dz}{dt}\end{bmatrix}$ is restricted with 2 equations concerned with estimated 3D flow $\frac{du}{dt}$,$\frac{dv}{dt}$, combining with the parallel restrictions, we can derive the velocity for each triangle, with pixel velocity bilinear interpolated on the center of the projected position.

After 3D velocity is calculated, we sample $4$ points on each triangular face to mimic as material points. Then we step one time interval to have the updated positions of each point. Since some of the points may move out of their initial position, we have to re-project the points back onto the fixed mesh surface. The projection is calculated as the minimal position, which is 
\begin{align}
	\begin{bmatrix}
		x_{n+1} \\ y_{n+1} \\z_{n+1}
	\end{bmatrix} &= \begin{bmatrix}
			x_{a} & x_{b} & x_{c} \\
			y_{a} & y_{b} & y_{c} \\
			z_{a} & z_{b} & z_{c}
	\end{bmatrix}\begin{bmatrix}
		\lambda_{0} \\ \lambda_{1} \\ \lambda_{2}
	\end{bmatrix} \nonumber \\
	\begin{bmatrix}
		\lambda_{0} \\ \lambda_{1} \\ \lambda_{2}
	\end{bmatrix} &= \min_{\Delta abc}\min_{\sum\lambda_{i}=1,\lambda_{i}\ge 0}\nonumber\\
		&\left\|\begin{bmatrix}
			x_{n}+dx \\ y_{n}+dy \\z_{n}+dz
		\end{bmatrix} - \begin{bmatrix}
			x_{a} & x_{b} & x_{c} \\
			y_{a} & y_{b} & y_{c} \\
			z_{a} & z_{b} & z_{c}
		\end{bmatrix}\begin{bmatrix}
		\lambda_{0} \\ \lambda_{1} \\ \lambda_{2}
		\end{bmatrix}\right\|
\end{align}
where we search for all $\Delta abc$ to find the nearest projection in the face to the updated points, formulated by a positive homogeneous combination coefficients $\lambda_{0},\lambda_{1},\lambda_{2}$.

After the 3D locations of material points were updated, we calculated the velocity on each face as the average of each material point. Then we calculate the pressure on each vertex position as the divergence of the velocity as
\begin{align}
	v_{\Delta abc} &= 
		\begin{bmatrix}
			x_{b}-x_{a} & x_{c}-x_{a} \\
			y_{b}-y_{a} & y_{c}-y_{a} \\
			z_{b}-z_{a} & z_{c}-z_{a}
		\end{bmatrix}\begin{bmatrix}
			\mu_{\Delta abc} \\ \lambda_{\Delta abc}
		\end{bmatrix} \nonumber \\
	p_{a} &= \text{div} v_{a} \nonumber \\
	&= \sum_{\Delta abc} \frac{\mu_{\Delta abc} + \lambda_{\Delta abc}}
		{S_{\Delta abc}}
\end{align}
where we express the velocity $v$ as the linear combination of triangle's edges and the coefficients are then added for all the triangles adjacent to a certain vertex $a$ to calculate the divergence. $S_{\Delta abc}$ represents the area of the triangle. The formula is consistent for discrete Gauss theorem
$\sum \text{div} v_{a} = \sum v_{\Delta abc} \cdot \vec{x}_{bc}$.

After vertex pressure is calculated, we apply the advection procedure to assure the velocity field is of no divergence, subscribing to the gradient operator of the pressure, which is to solve the equation.
\begin{align}
	\nabla p_{abc} &= \begin{bmatrix}
		\mu_{p} & \lambda_{p}
		\end{bmatrix}
		\begin{bmatrix}
			x_{b}-x_{a} & x_{c}-x_{a} \\
			y_{b}-y_{a} & y_{c}-y_{a} \\
			z_{b}-z_{a} & z_{c}-z_{a}
		\end{bmatrix} \nonumber \\
	\begin{bmatrix} p_{a} \\ p_{b} \\ p_{c} \end{bmatrix} &=  \nabla p_{abc} \cdot
	\begin{bmatrix}
		x_{a} & x_{b} & x_{c} \\
		y_{a} & y_{b} & y_{c} \\
		z_{a} & z_{b} & z_{c}
	\end{bmatrix} + \vec{c_{p}}
\end{align}
The pressure value on each vertex is estimated as a linear function of their position, with linear coefficients parallel to the triangle face, we can solve the parameter $\mu_{p},\lambda_{p}$ to get the gradient value for that face, then the velocity is subscribed with this velocity as what we do in 2D projection. 
\end{appendices}

\end{document}